%% file: main.tex
\documentclass[10pt,twocolumn,letterpaper]{article}
\PassOptionsToPackage{table}{xcolor}

\usepackage{cvpr}              

\usepackage[accsupp]{axessibility}  

\usepackage{algorithm}
\usepackage{algorithmicx}
\usepackage{algpseudocode}
\usepackage{subcaption}
\usepackage{float}


\definecolor{cvprblue}{rgb}{0.21,0.49,0.74}
\usepackage[pagebackref,breaklinks,colorlinks,allcolors=cvprblue]{hyperref}
\usepackage{multirow}
\usepackage{makecell}

\usepackage{graphicx}
\usepackage{booktabs}
\usepackage{adjustbox}

\def\paperID{10481} 
\def\confName{CVPR}
\def\confYear{2025}

\title{TopV: Compatible Token Pruning with Inference Time Optimization for Fast and Low-Memory Multimodal Vision Language Model}

\author{
 \textbf{Cheng Yang\textsuperscript{1*}},
 \textbf{Yang Sui\textsuperscript{2*$\dagger$}},
 \textbf{Jinqi Xiao\textsuperscript{1}},
 \textbf{Lingyi Huang\textsuperscript{1}},
 \textbf{Yu Gong\textsuperscript{1}},
 \textbf{Chendi Li\textsuperscript{3}},
 \\
 \textbf{Jinghua Yan\textsuperscript{3}},
 \textbf{Yu Bai\textsuperscript{4}},
 \textbf{Ponnuswamy Sadayappan\textsuperscript{3}},
 \textbf{Xia Hu\textsuperscript{2}},
 \textbf{Bo Yuan\textsuperscript{1}}
\\
 \textsuperscript{1}Rutgers University,
 \textsuperscript{2}Rice University,
 \textsuperscript{3}The University of Utah,
 \textsuperscript{4}California State University, Fullerton
\\
 \small{
 \texttt{cheng.yang@rutgers.edu, yang.sui@rice.edu, bo.yuan@soe.rutgers.edu}
 }
}

\begin{document}
\maketitle

\input{sec/0_abstract}   
\def\thefootnote{*}\footnotetext{Equal Contribution.}
\def\thefootnote{$\dagger$}\footnotetext{Corresponding Author.}

\input{sec/1_intro}
\input{sec/2_related}

\input{sec/3_method}
\input{sec/4_exp}

\input{sec/5_conclusion}

\noindent \paragraph{Acknowledgment} This work was supported in part by the NSF through award 2217154, NSF IIS-2224843, ITE-2429680, CCF-2239945, and CMMI-2152908.

\clearpage

{
    \small
    \bibliographystyle{ieeenat_fullname}
    \bibliography{main}
}

\input{sec/X_suppl}


\end{document}

%% file: sec/0_abstract.tex
\begin{abstract}
Vision-Language Models (VLMs) demand substantial computational resources during inference, largely due to the extensive visual input tokens for representing visual information. Previous studies have noted that visual tokens tend to receive less attention than text tokens, suggesting their lower importance during inference and potential for pruning. However, their methods encounter several challenges: reliance on greedy heuristic criteria for token importance and incompatibility with FlashAttention and KV cache. To address these issues, we introduce \textbf{TopV}, a compatible \textbf{TO}ken \textbf{P}runing with inference Time Optimization for fast and low-memory \textbf{V}LM, achieving efficient pruning without additional training or fine-tuning. Instead of relying on attention scores, we formulate token pruning as an optimization problem, accurately identifying important visual tokens while remaining compatible with FlashAttention. Additionally, since we only perform this pruning once during the prefilling stage, it effectively reduces KV cache size. Our optimization framework incorporates a visual-aware cost function considering factors such as Feature Similarity, Relative Spatial Distance, and Absolute Central Distance, to measure the importance of each source visual token, enabling effective pruning of low-importance tokens. Extensive experiments demonstrate that our method outperforms previous token pruning methods, validating the effectiveness and efficiency of our approach.


\end{abstract}

%% file: sec/1_intro.tex
\section{Introduction}
\label{sec:intro}

Vision-Language Models (VLMs) have significantly advanced the interpretation and synthesis of information across textual and visual modalities with their cutting-edge techniques. VLMs process a combination of visual inputs and textual data, such as an image accompanied by instructional text, to generate textual outputs like descriptions or answers to visual queries, facilitating a range of applications, including visual recognition \cite{menon2022visual, zhang2021vinvl, yuan2022volo, wu2024transferring}, image captioning \cite{bordes2024introduction, devlin2015language, zhou2020unified, mokady2021clipcap}, visual question answering \cite{antol2015vqa, jang2017tgif, xu2017video, maaz2023video}, and optical character recognition (OCR) \cite{mori1999optical, li2023trocr, fujitake2024dtrocr, zhang2024vision} tasks. 

Despite their versatility, VLMs require significant computational resources and substantial memory during inference. In addition to processing text tokens (e.g., system tokens, instruction tokens), VLMs need to handle a considerable volume of visual tokens, which substantially increase both computational costs and memory usage. For example, in the OCR task, LLaVA-v1.5 \cite{liu2024visual} translates an image into 576 visual tokens, constituting up to 87\% of the total input sequences. Similarly, InternVL \cite{chen2024internvl, chen2024far, gao2024mini}, depending on image resolution, produces between 256 and 1792 visual tokens, accounting for up to 95\% of the total tokens. These inefficiencies present a serious obstacle to scaling VLMs in resourced-limited environments.


Recently, observed by work \cite{chen2024image}, visual tokens often receive less attention compared to text tokens, indicating their lower importance during the VLM inference. Consequently, visual token pruning~\cite{chen2024image,tao2024dycoke,lin2024boosting} is proposed to reduce redundant visual information. FastV \cite{chen2024image} dynamically prunes unimportant visual tokens based on their received attention score. LLaVA-PruMerge \cite{shang2024llava} retains the most critical visual tokens and then merges less important tokens into them. VTW \cite{lin2024boosting} prunes all vision tokens from certain layers. However, they face three drawbacks: 


\begin{itemize} 
    \item \textbf{Greedy Heuristic Criteria for Token Importance}. Previous works such as FastV \cite{chen2024image}, and LLaVa-PruMerge \cite{shang2024llava} primarily rely on attention scores to measure the importance, assuming higher scores indicate greater relevance. In Contrast, VTW \cite{lin2024boosting} prunes all visual tokens after certain layers without considering importance metrics. As a result, these methods fail to accurately capture each token's contribution to the VLM model.

    \item \textbf{Incompatiblity with FlashAttention}. FlashAttention \cite{dao2022flashattention} computes output tokens by fusing the softmax with the Value matrix multiplication that reduces memory access costs and accelerates training/inference. However, previous works FastV \cite{chen2024image} and LLaVa-PruMerge \cite{shang2024llava} utilize the attention scores for token importance, requiring explicit computation and negating FlashAttention's efficiency benefits.

    \item \textbf{Incompatiblity with KV Cache}. Previous works such as FastV \cite{chen2024image} dynamically prune tokens at each decoding step, requiring variable KV cache sizes. As a result, the full KV cache from the prefilling stage must be stored, negating memory savings.
    
\end{itemize}

To address above issues within a token pruning framework, we propose TopV, an optimization-based token pruning method for efficient VLM inference, compatible with KV cache and FlashAttention \cite{dao2022flashattention} without extra training. Instead of using attention scores, we optimize token selection based on their contribution to subsequent layers, ensuring accurate importance estimation. Our approach avoids additional attention computations, leveraging FlashAttention fully. Furthermore, we prune tokens once during prefilling, maintaining a reduced set throughout decoding, effectively minimizing KV cache size and memory usage. Our contributions are summarized as follows:
\begin{itemize}

    \item We propose a training-free token reduction method for VLMs that optimizes pruning during the inference prefilling stage, reducing computation while maintaining compatibility with FlashAttention and KV cache. By pruning once and avoiding explicit attention score calculations, our approach enables a fast, low-memory VLM.

    \item We formulate the optimization problem by incorporating vision-specific factors such as feature similarity, spatial relationships, and central distance to enhance token selection and the visual-aware cost function. Token importance is then computed using the Sinkhorn algorithm, followed by a token recovery strategy after pruning.

    \item We conduct extensive experiments on LLaVA and InternVL2 models across multiple tasks, demonstrating that our method surpasses state-of-the-art token pruning approaches. Specifically, on InternVL2 models, it improves accuracy by 1.2\%, reduces visual FLOPs by 47\%, lowers dynamic memory usage by 61\%, and achieves 2.1× inference efficiency. For LLaVA models, it improves performance by 0.39\%, reduces visual FLOPs by 50\%, lowers dynamic memory usage by 49\%, and achieves 1.68× inference efficiency compared to previous methods.
    
\end{itemize}

%% file: sec/2_related.tex
\begin{figure*}[t]
  \includegraphics[width=2\columnwidth]{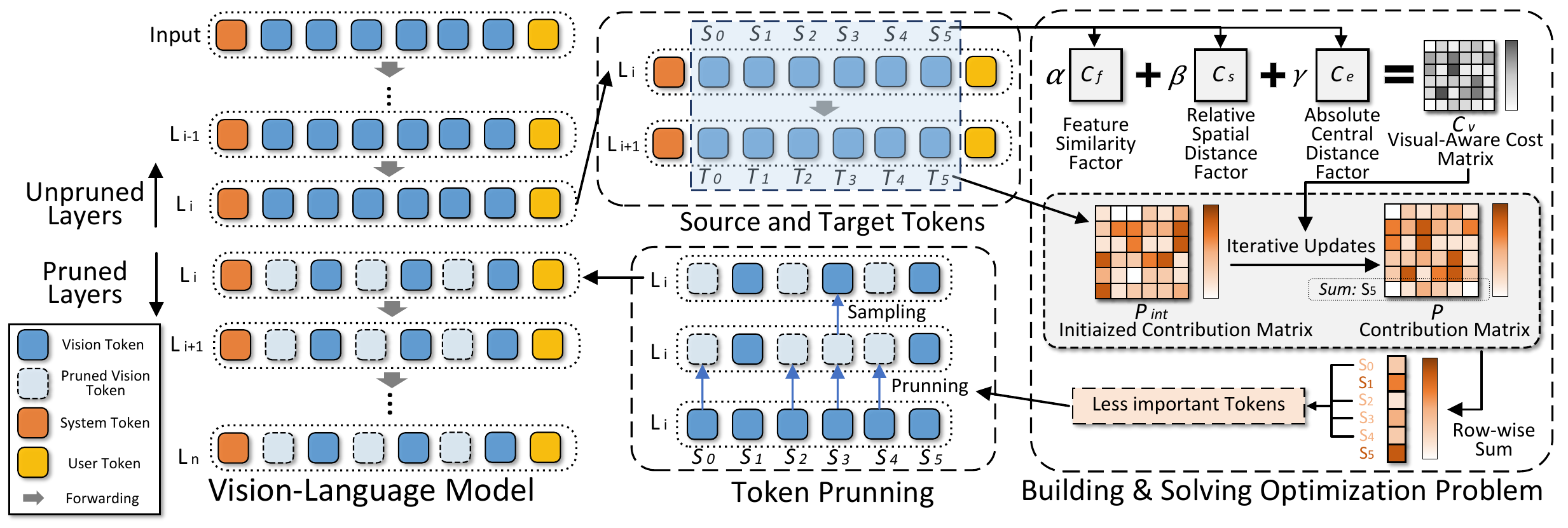}
  \caption{The pipeline of proposed TopV. In the prefilling stage during the inference time, we aim to prune the visual tokens in the $L_i$ layer. First, we select input visual tokens as source tokens and then collect output tokens after the Post-LN layer within the same layer as target tokens. We then formulate the token pruning as an optimization problem, incorporating factors such as feature similarity, relative spatial distance, and absolute central distance. Using the Sinkhorn algorithm to solve this problem, we obtain the contribution matrix of source tokens and their importance. Based on these values, we prune unimportant tokens and uniformly recover a subset of pruned tokens to maintain structural integrity. Starting from the layer $L_{i+1}$, tokens are consistently pruned, leading to faster, low-memory VLM inference. Following \cite{chen2024image}, we set $L_i=2$ in our experiments. Notably, our optimization process requires only 2 ms, comprising less than $1\%$ of total inference time.} 
  \label{fig:pipeline}
  \vspace{-2mm}
\end{figure*}

\section{Related Work} 
\label{sec:related}



\noindent\textbf{Vision-Language Models.} Recent advances in Vision-Language Models (VLMs) have made significant strides in integrating vision, video, and language modalities, with models like CLIP \cite{radford2021learning}, LLaVA \cite{liu2024improved, liu2024visual}, and BLIP-2 \cite{li2023blip} enabling tasks such as image captioning and zero-shot image-to-text generation. Video-LLaMA \cite{zhang2023video} and LLaVA-NeXT \cite{liu2024llavanext} expanded these capabilities to include video and audio processing, further enhancing multimodal understanding. However, as models scale to higher resolutions or multiple video frames, they require an exponentially growing number of tokens, leading to substantial computational challenges. For instance, LLaVA processes high-resolution images into thousands of tokens, and video models like VideoLLaVA \cite{lin2023video} allocate even more to represent multiple frames. This growing token demand presents a significant bottleneck, highlighting the need for more efficient computational strategies to fully realize the potential of VLMs.

\noindent\textbf{Visual Token Pruning for VLMs.} Visual tokens often vastly outnumber text tokens in VLMs. In the previous works, vision token pruning can be divided into two primary categories: pruning before the LLM and pruning within the LLM. In the case of pruning before the LLM, visual tokens are discarded before entering the model, thereby reducing the computational burden on the LLM. For instance, LLaMA-VID \cite{li2025llama} employs a Q-Former \cite{li2023blip} to compress visual tokens with the assistance of context tokens before passing them to the LLM. Similarly, DeCo \cite{yao2024deco} utilizes adaptive pooling to downsample visual tokens before they are entered into the LLM. Both methods focus on early token reduction, improving efficiency by keeping only the most relevant visual information. For pruning within the LLM, the model can consider token interactions and make pruning decisions based on more informed features in layers of LLM. For example, LLavaVolta \cite{chen2024llavolta} proposes multi-stage training to gradually reduce the number of visual tokens, though achieving this in a train-free manner remains challenging. FastV \cite{chen2024image} predicts the importance of visual tokens in the early layers of the LLM and prunes tokens accordingly. However, FastV relies on dynamic pruning and explicit calculation of attention maps, which can complicate KV cache management and FlashAttention. Finally, VTW \cite{lin2024boosting} employs aggressively pruning visual tokens in the middle layers of the LLM. While this approach works well for recognition tasks, it is less effective for tasks like OCR, caption generation, or VQA, where more detailed visual information is required.


%% file: sec/3_method.tex

\vspace{-2mm}
\section{Method}
\label{sec:method}

In this section, we introduce our token pruning method. We begin with formulating the visual token pruning in VLMs as an optimization problem. Then, we detail the essential components required to develop and solve this problem. Finally, we present our proposed recovery-based pruning strategy.

\subsection{Token Importance Formulation}

Obtaining the importance of each visual token can be formulated as an optimization problem. Given the input tokens in the $L_i$ layer, token pruning aims to keep tokens that have the greatest impact on subsequent layers, while removing those with minimal influence. As shown in Fig. \ref{fig:pipeline}, we define the input visual tokens as source tokens $\boldsymbol{\mathcal{S}} = \{s_i \in \mathbb{R}^{d} \mid i = 1, 2, \dots, N\}$, while the output tokens from this layer are viewed as target tokens $\boldsymbol{\mathcal{T}} = \{t_j \in \mathbb{R}^{d} \mid j = 1, 2, \dots, N\}$, where $d$ represents the dimensionality of each token. The objective is to identify the important source tokens $\boldsymbol{\mathcal{S}}$ that most significantly contribute to constructing the target tokens. Inspired by Optimal Transport \cite{peyre2019computational, genevay2016stochastic, montesuma2024recent, villani2009optimal}, the contribution of each source token can be obtained by solving this Optimal Transport optimization problem:
\begin{gather}
\mathbf{P}^* = \mathop{\arg\min}\limits_{\mathbf{P}} \left( \sum_{i=1}^N \sum_{j=1}^N \mathbf{P}_{i,j} \mathbf{C}_v(s_i, t_j) \right), \label{eq:optimization} 
\end{gather} 
where $\mathbf{P}$ represents the transport plan. $\mathbf{P}^*$ denotes the optimal transport plan, also referred to as the \textit{Contribution Matrix}, reflecting the contribution scores of the source tokens in constructing the target tokens. $\mathbf{C}_v(\cdot)$ is the cost function representing the cost of transporting mass from $\boldsymbol{\mathcal{S}}$ to $\boldsymbol{\mathcal{T}}$.  After obtaining the $\mathbf{P}^*$, tokens with lower contributions can be identified and discarded to reduce computational cost and memory usage. Note that we follow \cite{chen2024image} to measure the importance of visual tokens at $L_i=2$ layer and remove the corresponding tokens within layers starting from it in our experiments. 


Formulating and solving Eq. \eqref{eq:optimization} requires several key components. First, in Sec. \ref{sec:position}, we introduce a token analysis to determine the position of the input tokens $\boldsymbol{\mathcal{S}}$ and target tokens $\boldsymbol{\mathcal{T}}$ within the layer. Second, in Sec. \ref{sec:cost}, we establish a visual-aware cost function $\mathbf{C}_v$ by incorporating the visual properties of the tokens in VLMs. Third, in Sec. \ref{sec:pruning}, we iteratively solve Eq. \eqref{eq:optimization} to obtain the \textit{Contribution Matrix} $\mathbf{P}^*$, and then unimportant tokens are able to be pruned according to their contribution scores. Finally, in Sec. \ref{sec:recover}, we employ a token recovery strategy to support and complement the collapsed visual tokens.

\subsection{Determining the Target Tokens}
\label{sec:position}

\begin{figure}[t]
  \includegraphics[width=1\columnwidth]{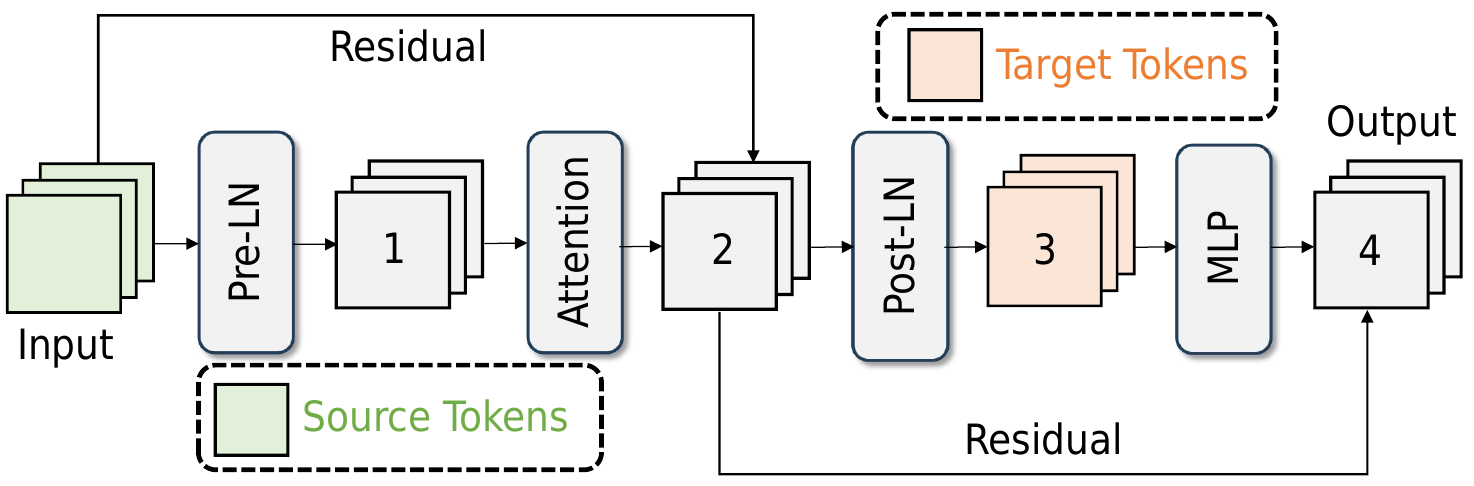}
  \caption{Illustration of target token positions within a transformer layer of VLMs. Positions 1, 2, 3, and 4 correspond to the outputs from the Pre-LN layer, Attention layer, Post-LN layer, and MLP layer, respectively. We empirically select the output after the Post-LN layer (Position 3) as the target tokens.}
  \label{fig:layer}
  \vspace{-2mm}
\end{figure}

Given the to-be-pruned visual tokens as source tokens $\boldsymbol{\mathcal{S}}$, we analyze to determine the appropriate position of target tokens $\boldsymbol{\mathcal{T}}$. Fig. \ref{fig:layer} illustrates a typical transformer block for LLMs within VLMs, where potential options for selecting target tokens include the outputs of: (1) Pre-LN, (2) Multi-Head Attention, (3) Post-LN, and (4) MLP.

First, since the core information processing primarily occurs in the Attention module, the Pre-LN output does not effectively capture the full function of the transformer block. Second, outputs after the Attention and MLP modules pass through the Residual module. The similarity induced by such residual connections can obscure distinctions between source and target tokens. By adopting the Post-LN output, we address this issue, ensuring that the target tokens reflect meaningful distinctions from the source tokens while also capturing the primary function of the Attention module.

Consequently, we select the output of (3) Post-LN as our target tokens. In Sec. ~\ref{sec:target}, we present ablation studies evaluating various target token positions, with empirical results supporting our analysis.

\subsection{Visual-Aware Cost Function}
\label{sec:cost}
In Eq. \eqref{eq:optimization}, the cost function is defined as $\mathbf{C}_v(\boldsymbol{\mathcal{S}}, \boldsymbol{\mathcal{T}}) \in \mathbb{R}^{N \times N}$ where each element $\mathbf{C}_v(s_i, t_j)$ quantifies the cost of moving a unit of mass from $s_i \in \boldsymbol{\mathcal{S}}$ to $t_j \in \boldsymbol{\mathcal{T}}$, where $i = 1, 2, \cdots, N, j = 1, 2, \cdots, N$. In the following section, we simply adopt $\mathbf{C}_v(s_i, t_j)$ to illustrate the proposed factors. 

\noindent\textbf{Feature Similarity Factor.} Transforming similar tokens requires minimal adjustment, which lowers the overall cost. Therefore, we incorporate the similarity between the source token $s_i \in \boldsymbol{\mathcal{S}}$ and target token $t_j \in \boldsymbol{\mathcal{T}}$ as the basic cost function. The similarity factor assigns a lower cost to pairs of tokens that are close in feature space, encouraging the optimal transport solution to align similar source and target tokens. Conversely, dissimilar pairs receive a higher cost, effectively penalizing less relevant matches. This similarity-based cost is effectively measured using the L2 norm:
\begin{equation}
    \mathbf{C}_f(s_i, t_j) = \left\|s_i - t_j\right\|_F^2,
    \label{eq:cost_dis}
\end{equation}
where each entry $\mathbf{C}_f(s_i, t_j)$ represents the transformation cost between a specific pair of tokens $(s_i, t_j)$.

\noindent\textbf{Relative Spatial Distance Factor.} In vision-related models (e.g., ViT, VLMs), considering spatial relationships among visual tokens is crucial for accurately capturing each token’s relative location and the overall structural layout within an image, For instance, positional encodings \cite{su2024roformer} is commonly used in CLIP \cite{radford2021learning}, ViT \cite{dosovitskiy2021an} and VLMs \cite{shang2024llava}, where positional information is encoded through transformations of token embeddings. These position encodings allow the tokens to maintain spatial correspondence with the original image. Generally, in vision tasks, the correlation between neighboring tokens is stronger than between tokens that are far apart, resulting in lower costs for transformations. To effectively leverage this spatial awareness, we introduce a Relative Spatial Distance Factor into the cost function. Specifically, we define the transport cost between spatially proximate tokens (i.e., tokens representing the same or neighboring regions within the image) to be lower than the cost for tokens representing distant regions. This spatial distance factor is defined by the relative Gaussian distance:
\begin{equation}
    \mathbf{C}_s(s_i, t_j) = {1 - \exp\left( -\frac{(x_{s_i} - x_{t_j})^2 + (y_{s_i} - y_{t_j})^2}{2\sigma^2} \right)},
    \label{eq:cost_p}
\end{equation}
where ($x_{s_i}$, $y_{s_i}$) and ($x_{t_j}$, $y_{t_j}$) represent the positions indices of the source token $s_i$ and the target token $t_j$. For the LLaVA-v1.5-7B model, which has 576 vision tokens, the original image is divided into 24 $\times$ 24 patches, meaning that the positional indices $x, y \in \{0, 1, \cdots, 23\}$. We illustrate the Gaussian distance from a source token $s_i$ to a target token $t_j$ used in relative spatial distance $\mathbf{C}_e(s_i, t_j)$. 

\noindent\textbf{Absolute Central Distance Factor.} Central parts of images typically carry the most relevant visual information for tasks such as object detection and scene understanding. Transforming source tokens from less informative areas, such as the edges of an image, into other areas demands extra information, leading to a higher cost. Accordingly, we introduce the Absolute Central Distance Factor into the cost function:
\begin{equation}
    \mathbf{C}_e(s_i, t_:) = \sqrt{(x_{s_i} - x_c)^2 + (y_{s_i} - y_c)^2},
    \label{eq:cost_edge}
\end{equation}
where $c_x$ and $c_y$ denote the coordinates of the center token in the original image. Specifically, $x_c = x_y = \sqrt{N}/2$, and $x, y \in \{0, 1, \cdots, \sqrt{N}\}$. Note that we only consider this factor for source tokens. Therefore, for a given source token $s_i$, the values of $\mathbf{C}_e(s_i, t_j)$ remain constant for all $j = 1, 2, \cdots, \sqrt{N}$. 
we depict the Euclidean distance from a source token $s_i$ to center coordinates used in the absolute central distance $\mathbf{C}_e(s_i, t_:)$.

\noindent\textbf{Overall Visual-Aware Cost Function.} We normalize each of the three factors to a range of $[0, 1]$ to avoid extreme imbalances among the terms. Then, taking into account these normalized factors, the visual-aware cost function is formulated as follows:
\begin{equation}
    \mathbf{C}_v(s_i, t_j) = \alpha \mathbf{C}_f(s_i, t_j) + \beta \mathbf{C}_s(s_i, t_j) + \gamma \mathbf{C}_e(s_i, t_j)
    \label{eq:cost},
\end{equation}
where $\alpha, \beta, \gamma$ serve as hyperparameters that adjust the relative weight of each cost factor. We assign a smaller value to $\gamma$ compared to $\alpha$ and $\beta$ to prevent excessive pruning of tokens at the image edges. More ablation studies are discussed in Sec. \ref{sec:ablation}.

\renewcommand{\algorithmicrequire}{\textbf{Inputs:}}
\renewcommand{\algorithmicensure}{\textbf{Outputs:}}
\begin{algorithm}[t]
\caption{Sinkhorn Algorithm for Token Pruning}
\begin{algorithmic}[1]
\Require Visual Tokens (Source Tokens) $\boldsymbol{\mathcal{S}}$, Target Tokens $\boldsymbol{\mathcal{T}}$, Visual-aware Cost Matrix $\mathbf{C}_v$, Temperature Parameter $\epsilon$, Max Iteration $T$, Tolerance $\delta$. 
\Ensure Contribution Matrix $\mathbf{P}^*$, Pruned Visual Tokens $\boldsymbol{\mathcal{S}}_p$.

\State $\boldsymbol{p}$, $\boldsymbol{q}$ $\gets$ \texttt{Norm}$(\boldsymbol{\mathcal{S}})$, \texttt{Norm}$(\boldsymbol{\mathcal{T}})$
\State $\boldsymbol{u}$, $\boldsymbol{v}$ $\gets$ $\mathbf{1}^{[N]}$,$ \mathbf{1}^{[N]}$
\State $\mathbf{K}$ $\gets$ $\exp(-\mathbf{C}_v / \epsilon)$
\State $t \gets 0$
\Repeat
    \State $t \gets t + 1$
    \State $\boldsymbol{v}_{t+1} \gets \boldsymbol{q} / (\mathbf{K}^T \exp(\boldsymbol{u}_{t} / \epsilon))$
    \State $\boldsymbol{u}_{t+1} \gets \boldsymbol{p} / (\mathbf{K} \exp(\boldsymbol{v}_{t+1} / \epsilon))$
    
\Until {(($\|\boldsymbol{u}_t - \boldsymbol{u}_{t+1}\| < \delta$) and ($\|\boldsymbol{v}_t - \boldsymbol{v}_{t+1}\| < \delta$)) \textbf{or} $t \geq T$}
\State $\mathbf{P}^*$ $\gets$ $\boldsymbol{u} \mathbf{K} \boldsymbol{v}$ \Comment{Contribution Matrix.}
\State $\mathbf{I} \gets \texttt{Sum}(\mathbf{P}^*, \texttt{dim}=1)$ \Comment{Token Importance.}
\State $\mathbf{I} \gets \texttt{Sort($\mathbf{I}$)}$
\State $\boldsymbol{\mathcal{S}}_{top} \gets \texttt{TopK} (\mathbf{I})$ \Comment{Important Tokens.}
\State $\text{Remove the unimportant visual tokens } \boldsymbol{\mathcal{S}}_p = \boldsymbol{\mathcal{S}} \setminus \boldsymbol{\mathcal{S}}_{top}$
\end{algorithmic}
\label{alg:ot}
\end{algorithm}

\subsection{Obtaining Token Importance}
\label{sec:pruning}

\textbf{Solving the Optimization Problem.} To solve Eq. \eqref{eq:optimization}, we employ the Sinkhorn algorithm \cite{knight2008sinkhorn, cuturi2013sinkhorn, pham2020unbalanced} to iteratively solve the optimization problem as shown in Alg.~\ref{alg:ot}. $\boldsymbol{p}$ and $\boldsymbol{q}$ are the normalized distributions over the source tokens $\boldsymbol{\mathcal{S}}$ and target tokens $\boldsymbol{\mathcal{T}}$, ensuring that the transport plan $\mathbf{P}$ satisfies the marginal constraints. The vector $\boldsymbol{u}$ and $\boldsymbol{v}$ are dual variables that iteratively scale the source and target tokens, adjusting the transport plan to match the distributions $\boldsymbol{p}$ and $\boldsymbol{q}$. During each iteration, $\boldsymbol{u}$ and $\boldsymbol{v}$ are updated to refine the transport plan $\mathbf{P}$, ultimately balancing the flow between source and target tokens according to these distributions. Empirically, the convergence is very fast for our scenarios, which only require three iterations. 


\noindent\textbf{Calculating the Importance of Source Token.}
After solving the problem, we obtain the Contribution Matrix $\mathbf{P}^* \in \mathbb{R}^{N \times N}$ where each element $\mathbf{P}^*_{ij}$ indicates the contribution of source token $s_i$ in constructing target token $t_j$. Then, we compute Token Importance $\mathbf{I}$ by performing the row-wise summation of $\mathbf{P}^*$, yielding $\mathbf{I}_i = \sum_{j=1}^N \mathbf{P}^*_{i,j}$ for $i = 1, 2, \cdots, N$ where $\mathbf{I} \in \mathbb{R}^{N \times 1}$. Each element in $\mathbf{I}$ represents the individual importance of a source token. To perform token pruning, we select the top $k$ tokens with the highest importance scores in $\mathbf{I}$ and discard the remaining less important tokens.

\subsection{Token Recovering}
\label{sec:recover}


Following the pruning process, we introduce an additional uniform sampling-based token recovery method to reintegrate certain tokens in a structured manner, helping to prevent the potential visual tokens collapse, particularly in tasks where the generated tokens need to focus on diverse regions of the input image, such as the OCR task. As shown in Fig. \ref{fig:pipeline}, this approach involves uniformly sampling a subset of pruned tokens to restore, maintaining a balanced representation, and reducing the risk of visual collapse.

\subsection{Overall Algorithm}
The overall pipeline of TopV is shown in Fig. \ref{fig:pipeline}. During the prefilling stage, we begin by selecting the input visual tokens at layer $L$. We then collect the output tokens after the Post-LN layer within the same layer as target tokens. Next, we formulate and solve the optimization problem to determine token importance. Based on this, we prune the unimportant tokens and selectively recover some pruned tokens in a uniform pattern. Starting from the layer $L_{i+1}$, the corresponding tokens are consistently pruned, resulting in faster, low-memory VLM inference. We follow \cite{chen2024image} to set $L_i=2$ in our experiments. Notably, our optimization process takes only 2 ms, accounting for less than $1\%$ of total inference time in most cases. 

%% file: sec/4_exp.tex
\begin{table*}[t]
\centering
\setlength{\tabcolsep}{3pt}
\caption{Performance comparisons in AI2D, SQA\_IMG, MMMU, MMBench tasks. \textit{Score} represents the performance of tasks, \textit{Mem.} indicates maximum memory usage (in GB) for corresponding tasks, and \textit{Tput.} denotes the token throughput, measured in tokens per second, calculated as the total inference time divided by the number of tokens generated. \textit{FLOPs Ratio} indicated the proportion of reduced computation relative to the original vision tokens across all tasks. \textit{Lat.} represents the whole inference latency for corresponding tasks, in minutes and seconds (mm'ss) format. For MMBench, both CN and EN metrics are considered.}
\scalebox{0.71}{
\begin{tabular}{ccccccccccccccccccc}
\toprule
\multirow{2}{*}{Model} & \multirow{2}{*}{Method} & \multirow{2}{*}{\makecell{FLOPs \\ Ratio $\uparrow$}} & \multicolumn{4}{c}{AI2D} & \multicolumn{4}{c}{SQA\_IMG} & \multicolumn{4}{c}{MMMU} & \multicolumn{4}{c}{MMBench}\\
\cline{4-19}
& & & Score$\uparrow$ & Mem.$\downarrow$ & Lat.$\downarrow$ & Tput.$\uparrow$ & Score$\uparrow$ & Mem.$\downarrow$ & Lat.$\downarrow$ & Tput.$\uparrow$ & Score$\uparrow$ & Mem.$\downarrow$ & Lat.$\downarrow$ & Tput.$\uparrow$ & Score$\uparrow$ & Mem.$\downarrow$ & Lat.$\downarrow$ & Tput.$\uparrow$ \\
\midrule
LLaVA-v1.5-7B & Baseline & 0 & 55.18 & 14.17 & 10'03 & 5.13 & 69.51 & 14.37 & 6'33 & 5.18 & 35.1 & 35.90 & 44'25 & 5.09 & 59.97 & 15.05 & 77'50 & 5.29 \\
\rowcolor{gray!20}
LLaVA-v1.5-7B & TopV & 35\%  & 55.41 & 13.98 & 9'25 & 6.07 & 69.56 & 14.3 & 6'02 & 6.18 & 35.4 & 35.9 & 41'52 & 5.76 & 60.42 & 14.9 & 72'36 & 6.4 \\
LLaVA-v1.5-7B & FastV & 47\% & 55.27 & 14.69 & 11'15 & 4.69 & 68.91 & 16.54 & 7'36 & 4.47 & 35.8 & 45.73 & 59'58 & 3.78 & 59.62 & 21.14 & 98'01 & 4.20 \\
\rowcolor{gray!20}
LLaVA-v1.5-7B & TopV & 51\%  & 55.31 & 13.87 & 8'30 & 6.12 & 69.61 & 14.17 & 5'32 & 6.27 & 35.1 & 35.45 & 39'30 & 5.82 & 59.65 & 14.77 & 67'02 & 6.46 \\
\midrule

LLaVA-v1.5-13B & Baseline & 0 & 59.29 & 28.61 & 9'43 & 5.29 & 72.93 & 28.92 & 6'15 & 5.3 & 35 & 49.13 & 42'51 & 5.26 & 65.46 & 30.06 & 74'38 & 5.43 \\
\rowcolor{gray!20}
LLaVA-v1.5-13B & TopV & 35\%  & 59.38 & 28.27 & 8'15 & 6.25 & 73.27 & 28.66 & 5'22  & 6.26 & 35.2 & 47.79 & 36'51  & 6.11 & 65.63 & 29.41 & 63'58 & 6.32 \\
LLaVA-v1.5-13B & FastV & 48\% & 58.91 & 29.39 & 10'09 & 5.07 & 73.12 & 29.51 & 7'02 & 4.78 & 34.3 & 75.32 & 58'58 & 3.83 & 64.95 & 40.12 & 86'48 & 4.63 \\
\rowcolor{gray!20}
LLaVA-v1.5-13B & TopV & 50\%  & 59.27 & 27.91 & 7'58 & 6.45 & 73.17 & 28.32 & 5'07 & 6.57 & 34.7 & 47.29 & 35'21 & 6.38 & 65.16 & 29.11 & 60'23 & 6.56 \\
\midrule

InternVL2-2B & Baseline & 0     & 72.67 & 6.39 & 9'10 & 5.61 & 94.2 & 6.63 & 6'07 & 5.5 & 34.56 & 13.72 & 39'57 & 5.25 & 67.6 & 6.78 & 86'57 & 12.1 \\
InternVL2-2B & FastV & 47\%     & 71.57 & 6.92 & 11'16 & 4.64 & 93.75 & 7.69 & 7'45 & 4.34 & 33.67 & 32.65 & 52'46 & 3.98 & 69.46 & 7.62 & 136'56 & 7.68 \\
\rowcolor{gray!20}
InternVL2-2B & TopV & 48\%      & 73.35 & 5.57 & 7'50 & 6.57 & 94.22 & 5.99 & 4'50 & 6.39 & 34.6 & 13.33 & 36'20 & 5.93 & 69.7 & 5.98 & 75'17 & 14.57 \\
\midrule

InternVL2-26B & Baseline & 0     & 83.13 & 56.79 & 41'57 & 2.46 & 97.47 & 57.61 & 28'43 & 2.34 & 47.11 & 69.02 & 173'47 & 2.24 & 80.89 & 57.02 & 342'27 & 4.24 \\
InternVL2-26B & FastV & 46\%     & 81.27 & 57.67 & 49'05 & 2.1 & 96.82 & 58.12 & 33'12 & 2.02 & 46.91 & 165.23 & 230'12 & 1.69 & 80.86 & 58.87 & 490'38 & 2.96 \\
\rowcolor{gray!20}
InternVL2-26B & TopV & 47\%      & 83.34 & 55.91 & 34'03 & 2.86 & 97.67 & 56.73 & 24'26 & 2.75 & 46.98 & 68.26 & 157'36 & 2.47 & 81.27 & 56.33 & 300'52 & 4.83 \\
\bottomrule

\end{tabular}
}
\label{tab:t1}
\end{table*}

\begin{table*}[t]
\centering
\setlength{\tabcolsep}{3pt}
\caption{Performance comparisons in POPE, Nocaps, OCRBench, OK-VQA tasks.}
\scalebox{0.71}{
\begin{tabular}{ccccccccccccccccccc}
\toprule
\multirow{2}{*}{Model} & \multirow{2}{*}{Method} & \multirow{2}{*}{\makecell{FLOPs \\ Ratio $\uparrow$}} & \multicolumn{4}{c}{POPE} & \multicolumn{4}{c}{Nocaps(CIDEr)} & \multicolumn{4}{c}{OCRBench} & \multicolumn{4}{c}{OK-VQA}\\
\cline{4-19}
& & & Score$\uparrow$ & Mem.$\downarrow$ & Lat.$\downarrow$ & Tput.$\uparrow$ & Score$\uparrow$ & Mem.$\downarrow$ & Lat.$\downarrow$ & Tput.$\uparrow$ & Score$\uparrow$ & Mem.$\downarrow$ & Lat.$\downarrow$ & Tput.$\uparrow$ & Score$\uparrow$ & Mem.$\downarrow$ & Lat.$\downarrow$ & Tput.$\uparrow$ \\
\midrule
LLaVA-v1.5-7B & Baseline & 0     & 85.81 & 13.96 & 25'59 & 5.77 & 105.67 & 13.93 & 172'30 & 22.10 & 31.3 & 14.00 & 15'05 & 22.67 & 53.38 & 13.98 & 19'42 & 9.54 \\
\rowcolor{gray!20}
LLaVA-v1.5-7B & TopV & 35\%      & 84.96 & 13.78 & 24'27 & 6.54 & 104.12 & 13.78 & 165'39 & 25.77 & 31.6 & 13.89 & 14'36 & 24.96 & 52.96 & 13.81 & 18'37 & 10.91 \\
LLaVA-v1.5-7B & FastV & 47\%     & 82.21 & 14.10 & 29'09 & 5.15 & 104.08 & 14.07 & 351'08 & 10.86 & 30.3 & 14.45 & 31'20 & 10.91 & 52.87 & 14.04 & 26'08 & 7.19 \\
\rowcolor{gray!20}
LLaVA-v1.5-7B & TopV & 51\%      & 84.21 & 13.7 & 22'16 & 6.59 & 103.52 & 13.7   & 158'01 & 25.81 & 31 & 13.77 & 14'03 & 25.01 & 52.64 & 13.76 & 17'20 & 10.97 \\
\midrule
LLaVA-v1.5-13B & Baseline & 0 & 86.01 & 28.53 & 24'26 & 6.14 & 109.37 & 28.79 & 160'32 & 22.98 & 33.6 & 28.97 & 14'34 & 24.29 & 58.2 & 28.39 & 18'59 & 10.08 \\
\rowcolor{gray!20}
LLaVA-v1.5-13B & TopV & 35\%  & 86.12 & 28.07 & 21'31 & 7.08 & 108.23 & 28.32 & 144'28  & 26.15 & 34.1 & 28.39 & 13'03  & 27.62 & 58.02 & 27.91 & 17'12 & 10.99 \\
LLaVA-v1.5-13B & FastV & 48\% & 84.96 & 29.02 & 28'56 & 5.18 & 107.71 & 29.37 & 314'29 & 12.12 & 32.9 & 29.57 & 31'06 & 10.99 & 58.02 & 28.89 & 25'19 & 7.43 \\
\rowcolor{gray!20}
LLaVA-v1.5-13B & TopV & 50\%  & 85.53 & 27.79 & 20'30 & 7.32 & 107.67 & 28.02 & 138'52 & 27.61 & 33.6 & 28.13 & 12'26 & 28.97 & 57.92 & 27.63 & 16'46 & 11.27 \\
\midrule
InternVL2-2B & Baseline & 0      & 88.39 & 6.25 & 33'05 & 4.53 & 60.34 & 6.23 & 150'10 & 30.41 & 75.6 & 6.26 & 8'30 & 30.83 & 43.59 & 6.28 & 20'24 & 9.46 \\
InternVL2-2B & FastV & 47\%      & 87.89 & 6.45 & 41'05 & 3.65 & 60.22 & 6.58 & 432'57 & 10.55 & 68.8 & 6.78 & 22'38 & 12.03 & 43.28 & 6.56 & 31'02 & 6.22 \\
\rowcolor{gray!20}
InternVL2-2B & TopV & 48\%       & 88.02 & 5.68 & 25'05 & 5.55 & 64.16 & 5.45 & 130'20 & 32.17 & 72.9 & 5.85 & 7'39 & 34.79 & 43.02 & 5.47 & 18'17 & 10.79 \\
\midrule
InternVL2-26B & Baseline & 0     & 87.98 & 56.63 & 157'40 & 1.92 & 92.76 & 56.74 & 371'01 & 8.98 & 77.9 & 56.91 & 23'43 & 12.18 & 48.41 & 56.93 & 89'48 & 3.19 \\
InternVL2-26B & FastV & 46\%     & 87.06 & 57.36 & 194'29 & 1.56 & 93.62 & 57.93 & 910'12 & 3.66 & 73.5 & 58.21 & 50'02 & 5.78 & 48.23 & 58.17 & 135'02 & 2.12 \\
\rowcolor{gray!20}
InternVL2-26B & TopV & 47\%      & 87.32 & 56.02 & 128'54 & 2.35 & 95.02 & 56.11 & 320'29 & 10.42 & 75.1 & 56.22 & 20'10 & 14.33 & 48.12 & 56.08 & 81'21 & 3.53 \\
\bottomrule
\end{tabular}
}
\label{tab:t2}
\end{table*}

\section{Experiments}
\label{sec:exp}

\subsection{Experimental Settings}
\noindent\textbf{Model Settings.} To demonstrate the effectiveness of our method, we conduct mainly experiments on LLaVA-v1.5-7B and InternVL2-2B. LLaVA-v1.5-7B includes a larger language model with 576 visual input tokens, while InternVL2-2B has a smaller language model with variable visual input tokens from 256 to 1792 due to the dynamic image augmentation. Additionally, we evaluate our method on larger-scale models such as LLaVA-13B and InternVL2-26B. 

\noindent\textbf{Evaluation and Datasets.} To evaluate the performance in a task-agnostic setting with zero-shot, we conduct the experiments on a variety of VLM tasks, such as recognition and reasoning tasks with AI2D \cite{kembhavi2016diagram}, SQA\_image \cite{lu2022learn}, MMMU \cite{yue2024mmmu}, MMBench \cite{liu2025mmbench}, POPE \cite{li2023evaluating}; image captioning tasks with Npcaps \cite{agrawal2019nocaps} dataset; visual question answering tasks with OK-VQA \cite{marino2019ok}; and OCR tasks with OCRBench \cite{liu2023hidden}.

\noindent\textbf{Implementation Details.} The experiments are conducted on A6000 GPU for LLaVA-v1.5-7B and InternVL2-2B model, A100 GPU for LLaVA-v1.5-13B and InternVL2-26B model. The evaluation is performed with the Lmms-Eval \cite{zhang2024lmms} library. We set $\sigma = 10$ in Eq. \eqref{eq:cost_p}. For LLaVA models, we set the hypermeters $\alpha$, $\beta$, and $\gamma$ to 1, 1, and 0.01, respectively. For InternVL2 models, we set $\alpha$, $\beta$, and $\gamma$ to 1, 1, and 0.1, respectively. We set the pruning ratio as $50\%$ and the sampling interval to 4 for token recovering, achieving a 35\% FLOPs reduction in overall vision tokens for LLaVA models. With a pruning ratio of $60\%$ and a sampling interval of 6, our setting results in reducing round 50\% visual token FLOPs reduction. For InternVL2 models, we set the pruning ratio as $70\%$, and the sampling interval to 3, resulting in reducing round 47\% FLOPs ratio for the visual tokens.

\begin{table*}[t]
\centering
\small
\setlength{\tabcolsep}{4pt}
\caption{Performance comparison with VTW in AI2D, SQA IMG, MMMU, MMBench, POPE, Nocaps, OCRBench, OK-VQA tasks.}
\scalebox{0.98}{
\begin{tabular}{ccccccccccc}
\toprule
Model & Method     & FLOPs Ratio     & AI2D  & SQA\_IMG  & MMMU   & MMBench   & POPE & Nocaps & OCRBench & OK-VQA\\
\midrule
LLaVA-v1.5-7B & VTW  & 50\%   & 55.41 &  69.56  & 36.3   & 59.63         & 85.92 & 58.37     & 5.1   & 39.88 \\
\rowcolor{gray!20}
LLaVA-v1.5-7B & TopV & 51\%   & 55.31 &  69.61  & 35.1   & 59.65         & 84.21 & 103.52    & 31    & 52.64\\
\bottomrule
\end{tabular}
}
\label{tab:vtw}
\end{table*}

\subsection{Main Results}
\noindent\textbf{Token Pruning Result on Various Tasks.} Tab. \ref{tab:t1} and \ref{tab:t2} present the performance of the models after applying the TopV framework, with metrics performance score, memory usage (Mem.), and token throughput (Tput.).  Notably, memory usage is divided into static and dynamic components. Static memory refers to the memory used for model parameters, which remains unaffected by token pruning. For instance, the static memory usage of LLaVA-v1.5-7B and InternVL2-2B is approximately 13.8GB and 4.5GB, respectively. Dynamic memory, on the other hand, represents the runtime memory increase and more accurately reflects the effectiveness of various token pruning methods. In addition to the overall memory usage reported in Tab. \ref{tab:t1}, we further analyze dynamic memory usage in our comparisons.

For the InternVL2 models, when reducing FLOPs by round 47\% for the vision tokens, the model showed almost no loss of accuracy compared to the baseline, while dynamic memory usage is reduced by 23.8\% and inference efficiency improved to 1.23 $\times$ on average and improved to 1.31 $\times$ for some recognition tasks. When compared to FastV, which reduces FLOPs by 47\%, InternVL2 models shows a 1.2 higher score on average, with dynamic memory usage decreasing by 61\% and inference efficiency improving by 2.1$\times$.

For the LLaVA models, reducing FLOPs by 35\% of the vision tokens results in negligible accuracy loss compared to the baseline, with inference efficiency improving to 1.08 $\times$ on average and improving to 1.12 $\times$ for some recognition tasks.  When reducing FLOPs round 50\% of the vision tokens, the model outperforms the FastV, which reduces FLOPs by 47\%, by an average of $0.39\%$ in accuracy. Additionally, dynamic memory usage is reduced by 49\%, and inference efficiency improves by 1.68$\times$.

\begin{figure}[t]
  \includegraphics[width=\columnwidth]{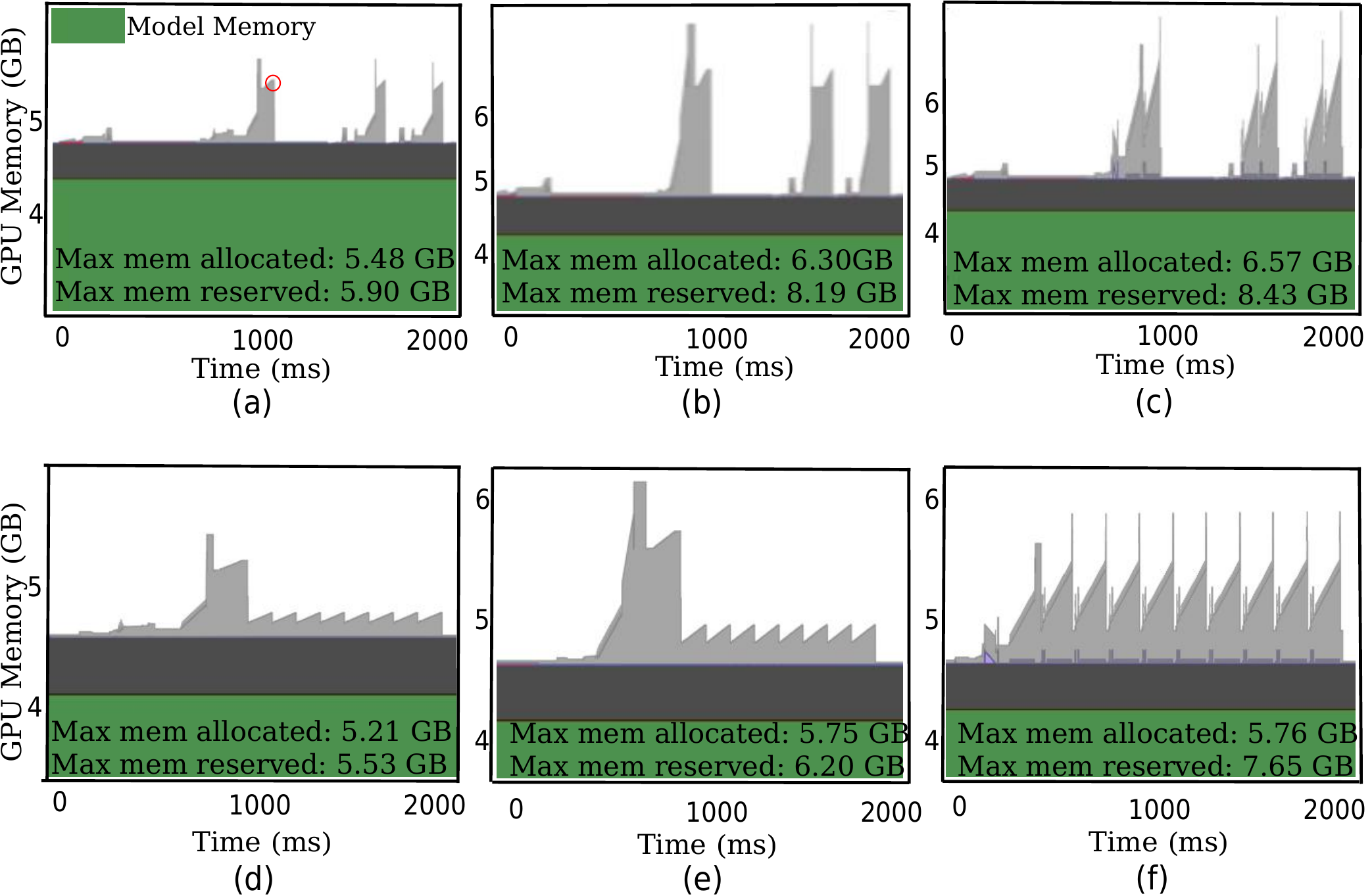}
  \caption{Memory Usage of TopV (a, d), Baseline (b, e), and FastV (c, f) on AI2D and OCRBench tasks for InternVL2-2B. The red circle indicates that a new token is currently being processed.} 
  \label{fig:mem}
\vspace{-4mm}
\end{figure}

\noindent\textbf{Memory Usage.} To further investigate the impact on memory usage resulting from compatibility with the KV cache, we conduct detailed profiling of memory usage on AI2D and OCRBench during the inference process.we select a few random samples for each task after GPU warmup (Fig. \ref{fig:mem}). Subfigures (a), (b), and (c) represent the memory usage of TopV, baseline, and FastV on the AI2D task, respectively, and (d), (e), (f) show the memory usage of these methods on the OCRBench task. In AI2D, a recognition task, memory is primarily allocated for the answer and end-of-text tokens (highlighted in Fig. \ref{fig:mem} (a)). With KV cache, memory from the first token is efficiently reused for the second, preventing extra allocations. In contrast, without KV cache (Fig. \ref{fig:mem} (c)), memory usage spikes again when generating the second token, indicating inefficient reuse. For OCR tasks, which generate multiple tokens, the absence of KV cache leads to continuous memory peaks, as each token requires additional allocation, increasing overall memory consumption.

Based on these observations, we assume that, although some intermediate values (e.g., tokens in VLMs) may not need to be retained during inference to save memory, when the remaining GPU memory is sufficient and the retained memory is not efficiently reusable, the CUDA backend opts for faster memory allocation strategies. This involves reallocating memory, rather than reusing the currently freed memory and then allocating new memory.  
It is important to note that we are discussing this issue in the context of a more general, off-the-shelf approach.

\begin{table}[t]
\centering
\small
\setlength{\tabcolsep}{15pt}
\caption{Impact of different target token positions within a block of the LLaVA-v1.5-7B model. Positions 1, 2, 3, and 4 represent the outputs from the Pre-LN layer, Attention layer, Post-LN layer, and MLP layer, respectively, as illustrated in Fig. \ref{fig:layer}. The "2*" represents the outputs from position 2 without the use of the residual module.}
\scalebox{0.8}{
\begin{tabular}{ccccc}
\toprule
Position & AI2D & SQA\_IMG & OK-VQA \\
\midrule
Baseline & 55.18 & 69.51 & 53.38 \\
1 & 54.17 & 68.32 & 47.28 \\
2* & 54.22 & 68.37 & 50.22 \\
2 & 54.27 & 68.86 & 48.99 \\
3 & \textbf{54.99} & \textbf{69.16} & \textbf{51.55} \\
4 & 53.21 & 66.41 & 43.98 \\
\bottomrule
\end{tabular}
}
\label{tab:target_layer}
\vspace{-2mm}
\end{table}

\begin{table}[t]
\centering
\small
\setlength{\tabcolsep}{10pt}
\caption{Evaluation Results on Video Question Answering Tasks.}
\vspace{-2mm}
\scalebox{0.62}{
\begin{tabular}{ccccccc}

\toprule
\multirow{2}{*}{Model} & \multirow{2}{*}{Method} & \multirow{2}{*}{\makecell{FLOPs \\ Ratio $\downarrow$}} & \multicolumn{2}{c}{TGIF} & \multicolumn{2}{c}{VideoMME(Overall)}  \\
\cline{4-7}
& & & Acc$\uparrow$ & Lat.$\downarrow$ & Acc$\uparrow$ & Lat.$\downarrow$ \\
\midrule
Video-LLaVA-7B& Baseline  & 0   & 0.27 &  414'13 & 0.3 & 41'20   \\
Video-LLaVA-7B & FastV  & 47\%   & 0.24 &  670'30 & 0.29 & 67'11  \\
\rowcolor{gray!20}
Video-LLaVA-7B & TopV  & 51\%   & 0.25 &  370'23 & 0.3 & 35'52  \\
\bottomrule
\end{tabular}
}
\label{tab:video}
\vspace{-4mm}
\end{table}

\noindent\textbf{Comparison with Vision Tokens Withdrawal in Middle Layer.} As shown in Tab. ~\ref{tab:vtw}, we also compared our method with the Vision Token Withdrawly \cite{lin2024boosting} approach, under the condition of reducing an equivalent proportion of vision tokens involved in computation. Unlike the token pruning method, VTW begins withdrawing all vision tokens from a block in the middle of the LLM. While this method is effective for tasks such as recognition questions, it performs poorly on tasks that require generating a large number of tokens, such as VQA, Caption, and OCR tasks. The reason is that, as the generated tokens evolve, the required vision features also change. However, in VTW's approach, once tokens are withdrawn from the middle, no vision tokens remain, leading to a loss of crucial vision information. 

\noindent\textbf{Comparison with FastV on Video Understanding Tasks.} As shown in Tab.~\ref{tab:video}, we compared our method with the FastV approach on video Understanding tasks. The results show that our method can outperform the FastV even with the higher visual token FLOPs reduction ratio.

\subsection{Ablation Studies}
\label{sec:ablation}

\label{sec:target}
\noindent\textbf{Impact of Outputs from Different Layers as Target Tokens.} As discussed in Sec~\ref{sec:method}, if a layer chosen for output is too early in the entire block, its output may not adequately represent the function of the current block. On the other hand, if the selected layer is too late, the presence of two residual connections may cause the output to be too similar to the source tokens. Here, we conduct the ablation studies only with the feature similarity factor. As shown in Tab. ~\ref{tab:target_layer}, selecting the output from Post-LN (3), as the target yields the best results. In contrast, using the MLP output as the output produces the poorest performance.

\noindent\textbf{Impact of Token Recovering.} To evaluate the effectiveness of our token recovery method, we conducted experiments to measure the impact on different tasks with Feature Similarity Cost. To achieve the FLOPs reductions of 35\% in LLaVA-v1.5-7B, we set the pruning ratio to $35\%$ when no token recovery is applied. When the token recovery is employed, the pruning ratio is set to 50$\%$. As shown in Table~\ref{tab:sampling}, token recovering has clear advantages in tasks that require generating a layer number of tokens to capture potentially useful tokens in subsequent steps.

\begin{table}[t]
\centering
\small
\setlength{\tabcolsep}{10pt}
\caption{Effect of Token Recovering. The results show the token pruning with and without the proposed token recovery method.}
\scalebox{0.95}{
\begin{tabular}{ccccccccc}
\toprule
Model & Method              & Sampling            & OCRBench \\
\midrule
LLaVA-v1.5-7B & Baseline     & -                       & 31.3 \\
LLaVA-v1.5-7B & TopV         & w/o                       & 29.7 \\
LLaVA-v1.5-7B & TopV         & w/                      & \textbf{30.5} \\
\midrule
InternVL2-2B  & Baseline     & -                      & 75.6 \\
InternVL2-2B  & TopV         & w/o                      & 70.5 \\
InternVL2-2B  & TopV         & w/                        & \textbf{71.3} \\
\bottomrule
\end{tabular}
}
\label{tab:sampling}
\end{table}

\noindent\textbf{Impact of Different Components of Visual-Aware Cost Function.} To validate the impact of each cost function in the Visual-Aware Cost Function with token recovering, we conducted experiments on four tasks. As shown in Tab. ~\ref{tab:cost}, the different cost functions are compatible and work together to enhance performance.  

\begin{figure}[t]
  \includegraphics[width=\columnwidth]{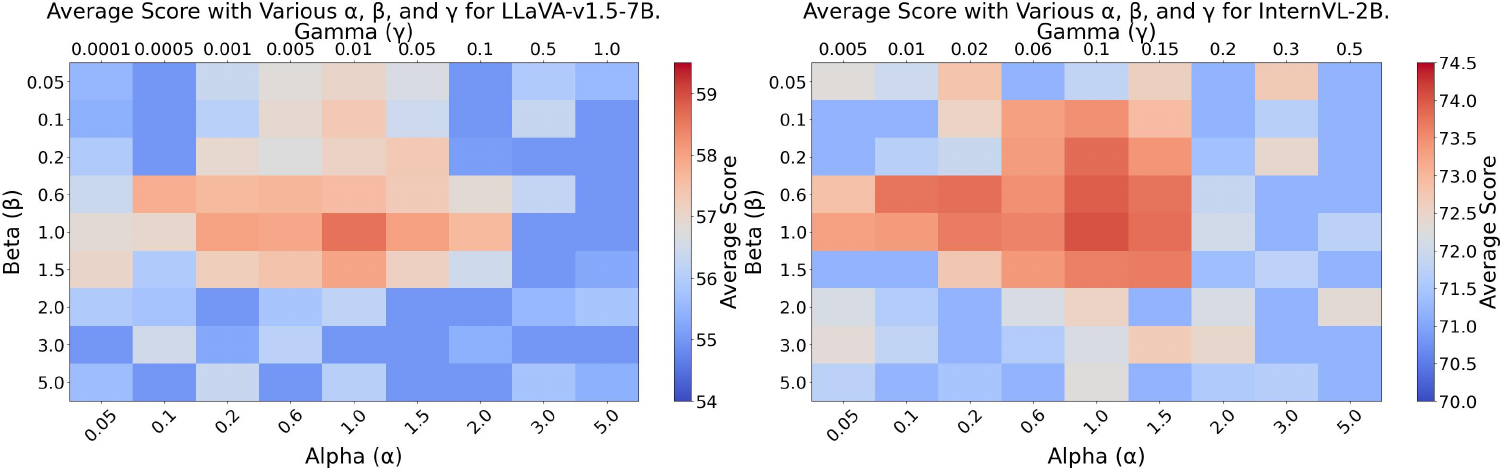}
  \caption{Two visualization heat maps for various hyperparameters.}
  \label{fig:heatmap}
  \vspace{-4mm}
\end{figure}

\noindent\textbf{The Sensitive to Various Hyperparameters.} With various $\alpha, \beta, \gamma$, we conduct the experiments and obtain the average score on AI2D, SQA\_IMG, POPE, OCRBench, and OK-VQA. As shown in Fig.~\ref{fig:heatmap}, the heatmaps for LLaVA-v1.5-7B (left) and InternVL2-2B (right) illustrate that while various hyperparameters affect performance, adjacent hyperparameter values yield similar scores, indicating that our method is not highly sensitive to them. 

\begin{table}[t]
\centering
\setlength{\tabcolsep}{1pt}
\caption{Impact of Different Components in Visual-Aware Cost Function. The 'F', 'S', and 'E' represent Feature Similarity, Relative Spatial Distance, and Absolute Distance Cost, respectively.}
\scalebox{0.82}{
\begin{tabular}{ccccccccc}
\toprule
Model & Method      & Costs         & AI2D  & SQA\_IMG  & MMMU      & OK-VQA \\
\midrule
LLaVA-v1.5-7B & Baseline & -         & 55.18 &  69.51    & 35.1    & 53.38 \\
LLaVA-v1.5-7B & TopV & F             & 54.99 &  69.16    & 35.1    & 51.55 \\
LLaVA-v1.5-7B & TopV & F+S           & 55.38 &  \textbf{69.61}    & 35.3    & 52.86 \\
LLaVA-v1.5-7B & TopV & F+S+E         & \textbf{55.41} &  69.56    & \textbf{35.4}    & \textbf{52.96} \\
\midrule
InternVL2-2B  & Baseline & -         & 72.67 &  94.2     & 34.56    & 43.59 \\
InternVL2-2B  & TopV & F             & 72.28 &  93.82    & 33.36    & 41.55 \\
InternVL2-2B  & TopV & F+S           & 73.32 &  94.12    & 34.36    & 42.97 \\
InternVL2-2B  & TopV & F+S+E         & \textbf{73.35} &  \textbf{94.22}    & \textbf{34.6}     & \textbf{43.02} \\
\bottomrule
\end{tabular}
}
\label{tab:cost}
\end{table}

\subsection{Visualization}
\begin{figure}[t]
   \centering
  \includegraphics[width=0.8\columnwidth]{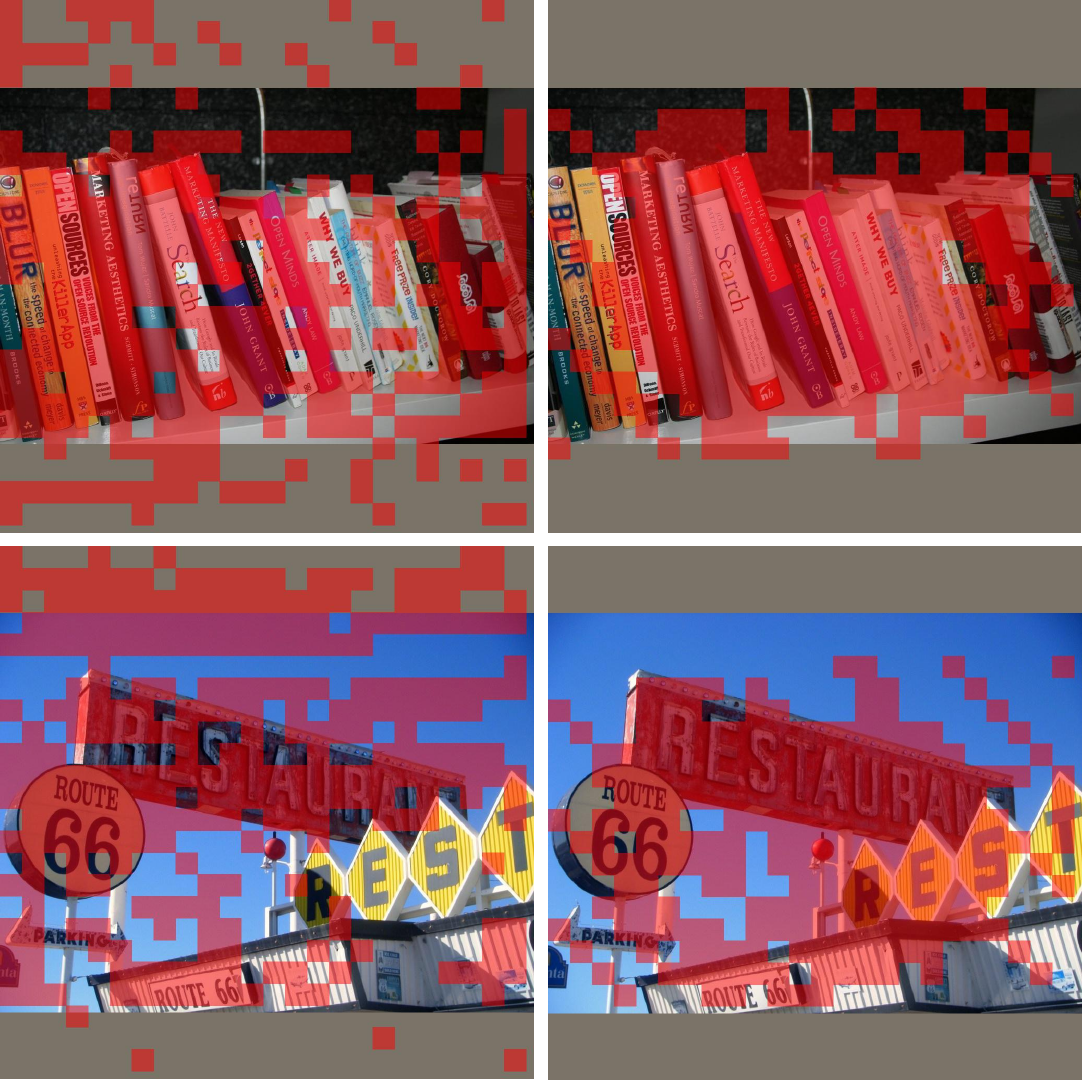}
  \caption{Two visualization examples of important tokens identified by various visual token pruning methods. \textbf{Left}: Important tokens selected by FastV.  \textbf{Right}: Important tokens selected by our TopV. The gray patches around the images are introduced during preprocessing, while the red patches indicate the token regions the model focuses on.}
  \label{fig:vis}
  \vspace{-4mm}
\end{figure}

\noindent\textbf{Important Token Distribution from various Pruning Method.} The important vision tokens selected by various pruning methods are visualized as shown in Fig. ~\ref{fig:vis}. Compared to FastV, TopV concentrates more tokens in the foreground region with red patches, avoiding unnecessary focus on the background or padded areas with gray patches, especially for surrounding padding areas.
  
%


%
%

%% file: sec/5_conclusion.tex
\section{Conclusion}
\label{sec:conclusion}

In this paper, we present TopV, a novel token pruning method designed to optimize inference efficiency in VLMs without additional training or fine-tuning. By formulating token pruning as an optimization problem, TopV effectively identifies and prunes redundant visual tokens, reducing computational demand and memory requirements. Our approach maintains compatibility with FlashAttention and reduces KV cache storage by performing pruning only once during the prefilling stage. Extensive experiments demonstrate that our method outperforms previous methods, validating the effectiveness and efficiency of our approach.

%% file: sec/X_suppl.tex
\clearpage
\setcounter{page}{1}
\maketitlesupplementary

\appendix
\section{Memory Analysis}
\label{sec:rationale}
\begin{figure}[ht]
  \includegraphics[width=1\columnwidth]{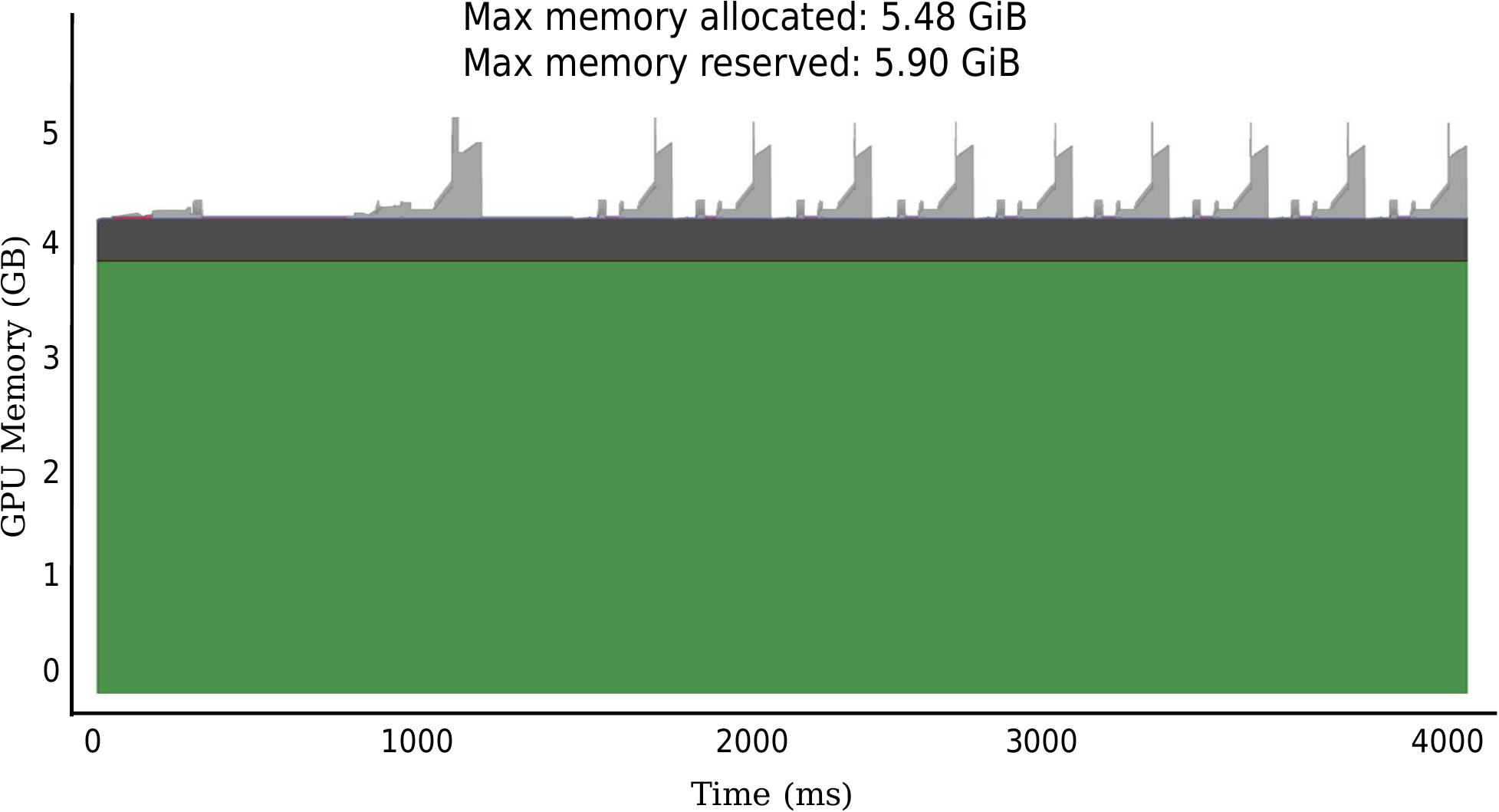}
  \caption{GPU Memory Usage of TopV on AI2D task for InternVL2-2B.} 
  \label{fig:yofo_ai2d}
  \vspace{-4mm}
\end{figure}

\begin{figure}[ht]
  \includegraphics[width=1\columnwidth]{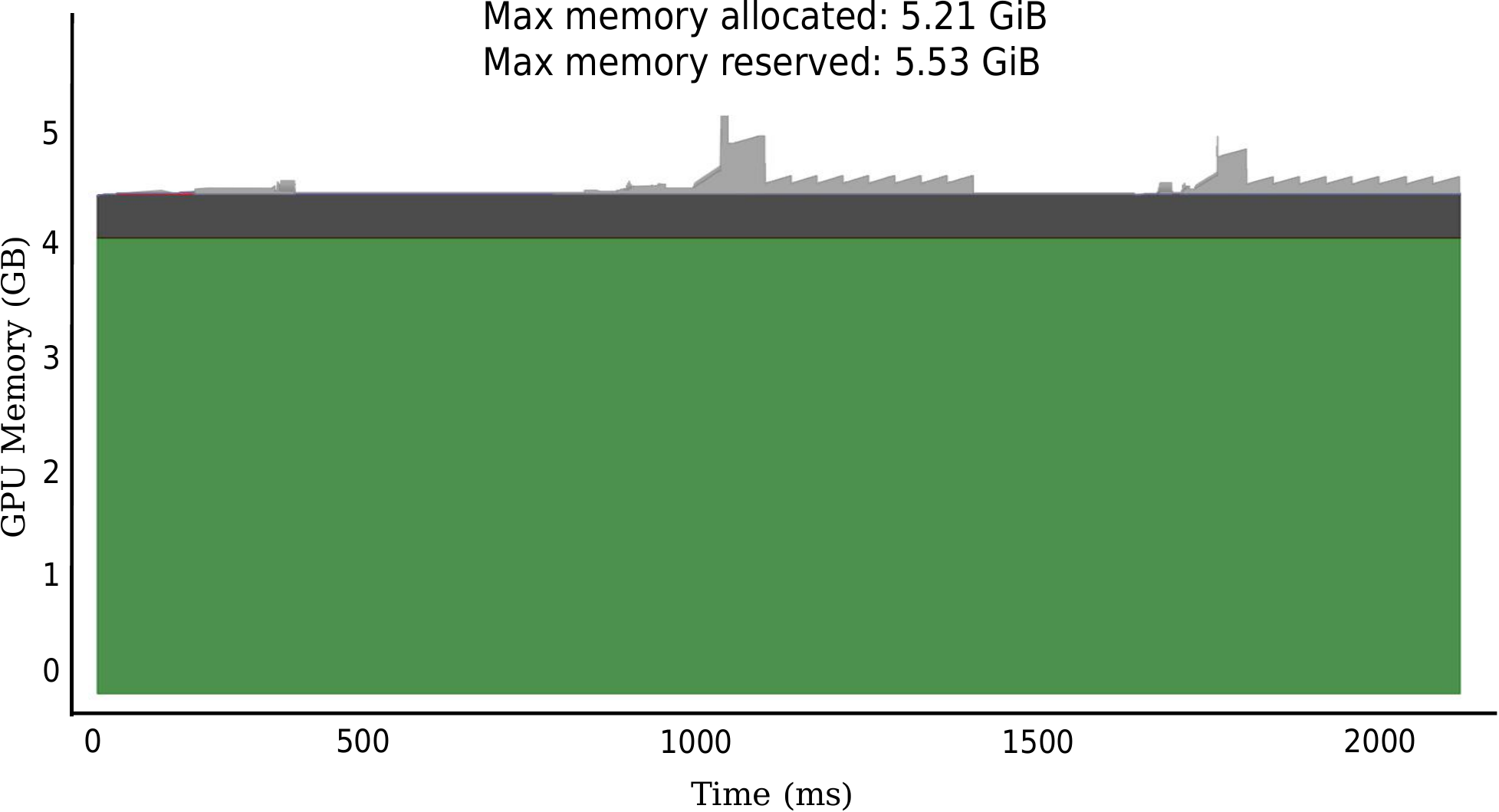}
  \caption{GPU Memory Usage of TopV on OCRBench task for InternVL2-2B.} 
  \label{fig:yofo_ocr}
  \vspace{-4mm}
\end{figure}

\begin{figure}[ht]
  \includegraphics[width=1\columnwidth]{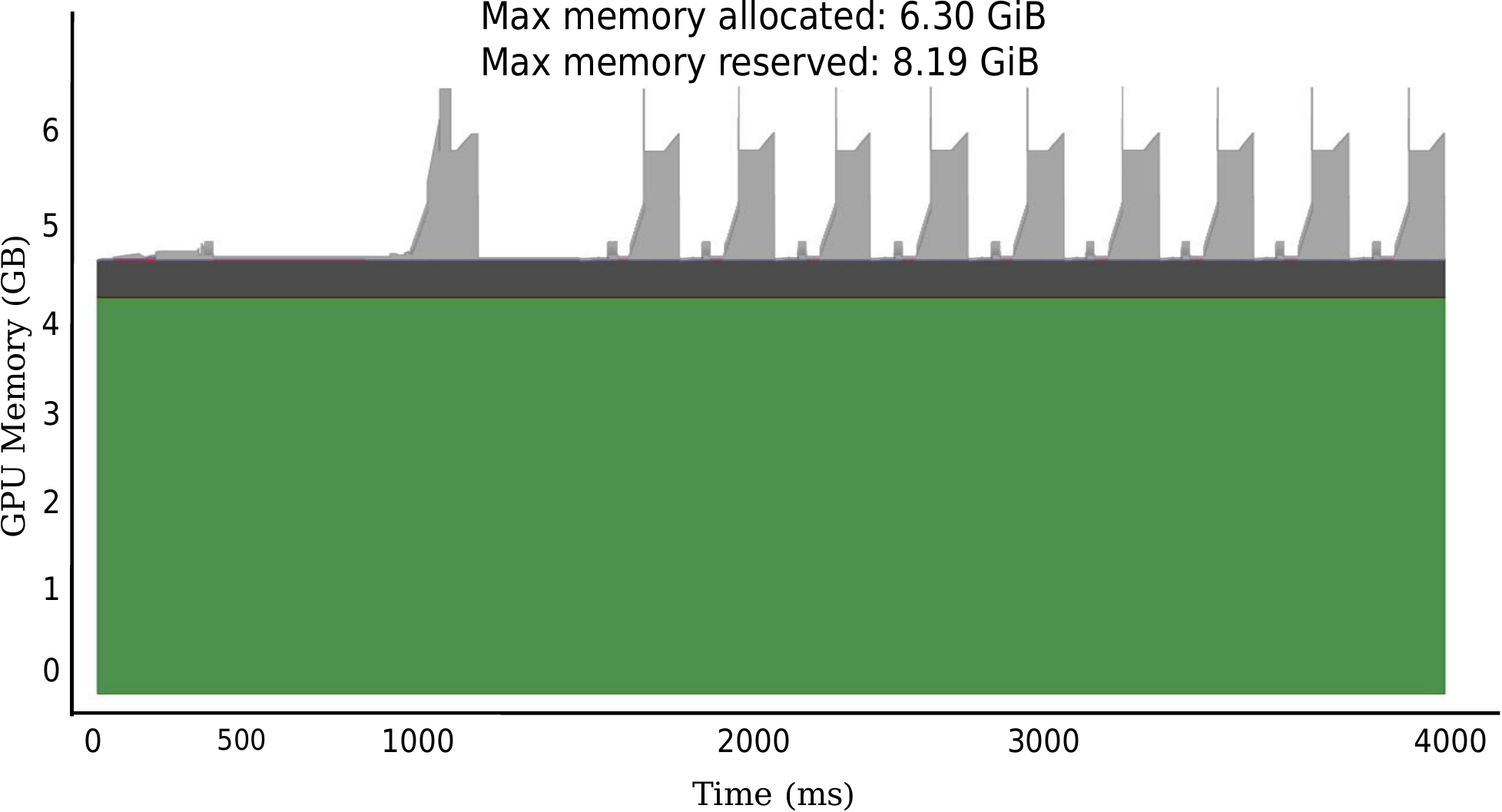}
  \caption{GPU Memory Usage of Baseline on AI2D task for InternVL2-2B.} 
  \label{fig:base_ai2d}
  \vspace{-4mm}
\end{figure}

\begin{figure}[ht]
  \includegraphics[width=1\columnwidth]{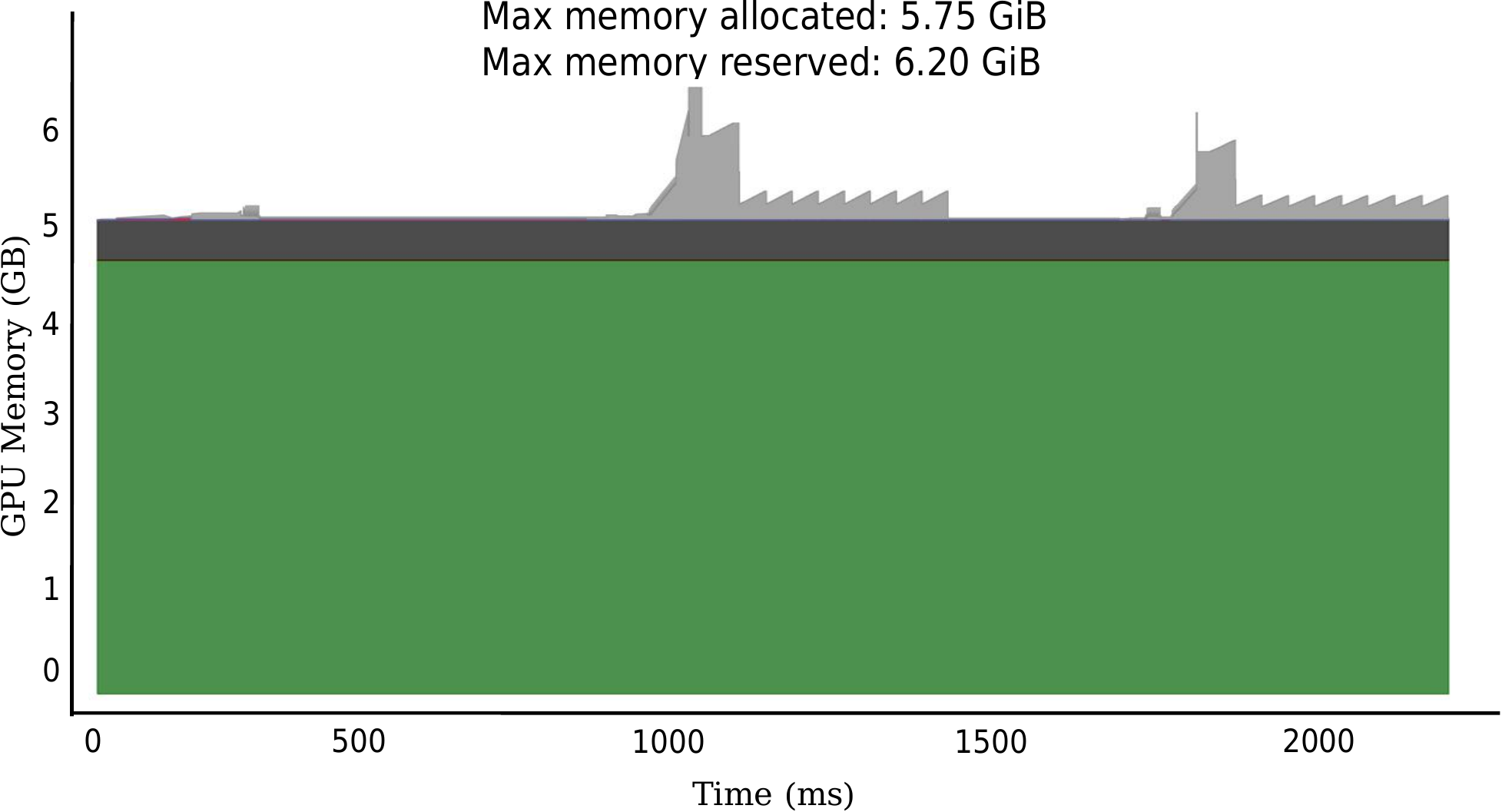}
  \caption{GPU Memory Usage of Baseline on OCRBench task for InternVL2-2B} 
  \label{fig:base_ocr}
  \vspace{-4mm}
\end{figure}

\begin{figure}[ht]
  \includegraphics[width=1\columnwidth]{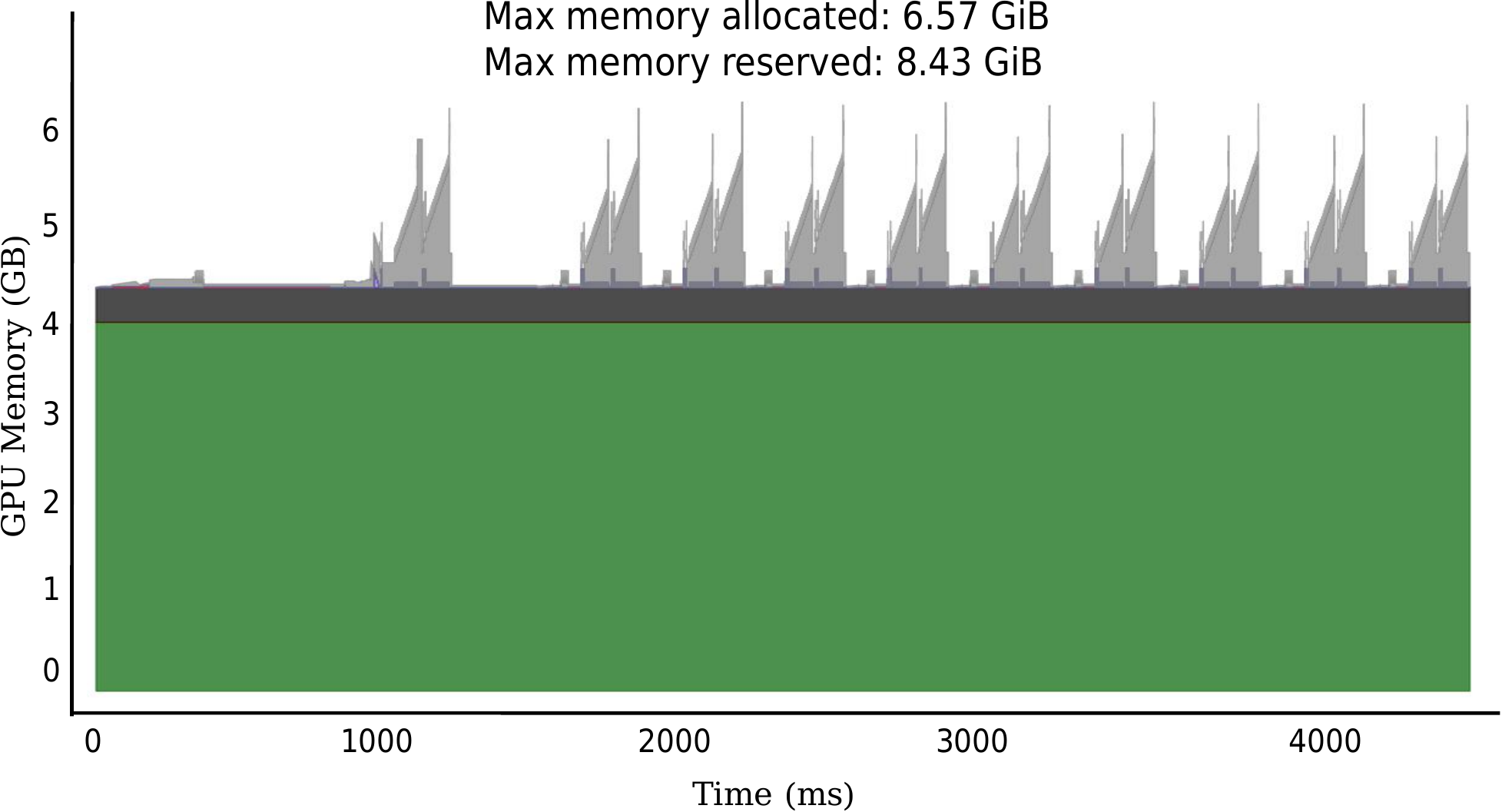}
  \caption{GPU Memory Usage of FastV on AI2D task for InternVL2-2B.} 
  \label{fig:fastv_ai2d}
  \vspace{-4mm}
\end{figure}

\begin{figure}[ht]
  \includegraphics[width=1\columnwidth]{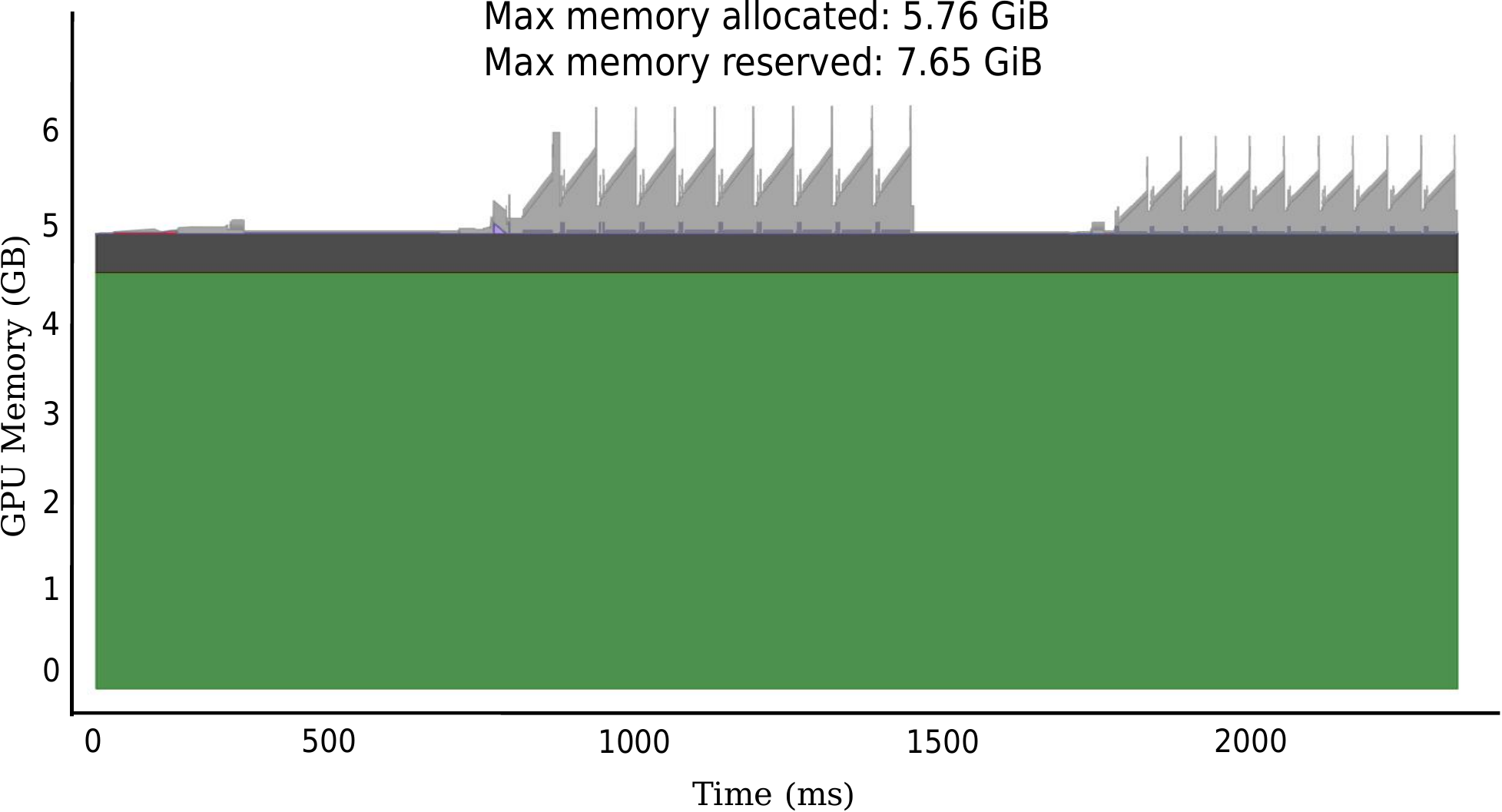}
  \caption{GPU Memory Usage of FastV on OCRBench task for InternVL2-2B.}
  \label{fig:fastv_ocr}
  \vspace{-4mm}
\end{figure}

Fig. ~\ref{fig:yofo_ai2d}-~\ref{fig:fastv_ocr} illustrate the GPU memory usage records of InternVL2-2B on the AI2D and OCRBench datasets, obtained using different methods. It is worth noting that conducting GPU Profiler analysis on all the tasks is challenging due to the substantial time and computational resources. We only performed GPU Profiler analysis after GPU warmup on a subset of the dataset. During the inference phase, we monitored the peak memory usage of GPU at the end of generating each token, with the peak memory usage for each dataset summarized in Tab. ~\ref{tab:t1} and ~\ref{tab:t2}. Specifically, Fig. ~\ref{fig:yofo_ai2d} and Fig. ~\ref{fig:yofo_ocr} show the results for TopV on the AI2D and OCRBench datasets, respectively; Fig. ~\ref{fig:base_ai2d} and Fig. ~\ref{fig:base_ocr} present the results for Baseline on AI2D and OCRBench; and Fig. ~\ref{fig:fastv_ai2d} and Fig. ~\ref{fig:fastv_ocr} display the results for the FastV on AI2D and OCRBench. Notably, TopV and FastV achieve a 48\% and 47\% reduction in vision token FLOPs.

In practical usage, the InternVL2-2B model typically consumes approximately 4.5GB of GPU memory, which is static and remains unaffected by token pruning techniques. This static memory allocation does not change with different pruning methods. However, pruning methods have distinct effects on dynamic memory usage. For instance, as illustrated in Fig. ~\ref{fig:yofo_ai2d}, ~\ref{fig:base_ai2d}, and ~\ref{fig:fastv_ai2d}, when performing inference on the same AI2D dataset, the dynamic memory usage for TopV, Baseline, and FastV is denoted as 0.98GB, 1.8GB, 2.07GB, respectively. Compared to Baseline and FastV, the TopV achieves dynamic memory savings of 45.6\% and 52.7\%. Similarly, as shown in Fig. ~\ref{fig:yofo_ocr}, ~\ref{fig:base_ocr}, and ~\ref{fig:fastv_ocr}, when inferring the same OCRBench dataset, the dynamic memory usage for TopV, Baseline, and FastV is 0.71GB, 1.25GB, 1.26GB, respectively. In this case, the TopV method results in dynamic memory savings of 43.2\% and 43.7\% relative to Baseline and FastV, respectively.

\section{Discussion}

Orthogonal to token pruning, memory efficiency, model compression ~\cite{frantar2022gptq,lin2024awq,yang2024moe, sui2021chip,sui2024co,xiao2023haloc,xiao2023comcat,xiao2024coap,sui2024bitsfusion} and KV cache compression~\cite{liu2024kivi,zhang2023h2o} are another promising techniques to enhance inference speed and reduce memory usage. An important question for future research is how to effectively balance token pruning and model compression/KV cache compression, potentially leading to extremely efficient multimodal language models.

\section{Evaluation Examples}

Tab. \ref{tab:ocr}, \ref{tab:caption}, and \ref{tab:writing} provide additional results for the dialogue tasks. While maintaining performance, we reduced the vision token FLOPs by 51\%, and 48\% for the LLaVA-v1.5-7B and InternVL2-2B models, respectively. As shown in Tab. \ref{tab:ocr}, \ref{tab:caption}, \ref{tab:writing}, which correspond to the OCR, Captioning, and Writing tasks, it is clear that applying our TopV method yields results that are closer to the ground truth.

In the OCR task presented in Tab. \ref{tab:ocr}, the model needs to recognize the word "Jump". The TopV method, applied to the LLaVA-v1.5-7B model, successfully identifies the word. while the Baseline fails to recognize it, and FastV only detects the letter "J". For the Writing task, as detailed in Tab. \ref{tab:writing}, the model is tasked with providing a comprehensive description of the image content. Both the TopV and Baseline generate similar descriptions, offering a more detailed description than FastV. In the Captioning task, TopV correctly identifies the rabbit, whereas both the Baseline and FastV methods erroneously classify it as a baby.

\begin{table}[t]
\centering
\setlength{\tabcolsep}{10pt}
\caption{Comparison of TopV, Baseline, and FastV in the OCR Task for LLaVA-v1.5-7B.}
\begin{tabular}{@{}p{0.4\textwidth}@{}}
\toprule
\textbf{Visual input example, OCR Task:} \\
\midrule
\end{tabular}
\begin{tabular}{@{}c@{}}
\adjustbox{valign=t}{\includegraphics[width=0.2\textwidth]{images/examples/ocr_869_cropped.pdf}}
\end{tabular}
\begin{tabular}{@{}p{0.4\textwidth}@{}}
\textbf{User:} what is written in the image? \\
\midrule
\textbf{TopV:} The image features a handwritten signature, which appears to be a cursive "\textbf{\color{green}{Jump}}" written in black ink. \\
\midrule
\textbf{Baseline:} The image features a handwritten signature, which appears to be a cursive or script-style writing. \\
\midrule
\textbf{FastV:} The image features a handwritten signature, which is a cursive letter "\textbf{\color{red}{J}}" written in black ink. \\
\bottomrule
\end{tabular}
\label{tab:ocr}
\end{table}


\begin{table}[htbp]
\centering
\caption{Comparison of TopV, Baseline, and FastV in the Captioning Task for InternVL2-2B.}
\begin{tabular}{@{}p{0.4\textwidth}@{}}
\toprule
\textbf{Visual input example, Captioning Task:} \\
\midrule
\end{tabular}
\vspace{2mm} 
\begin{tabular}{@{}c@{}}
\adjustbox{valign=t}{\includegraphics[width=0.2\textwidth]{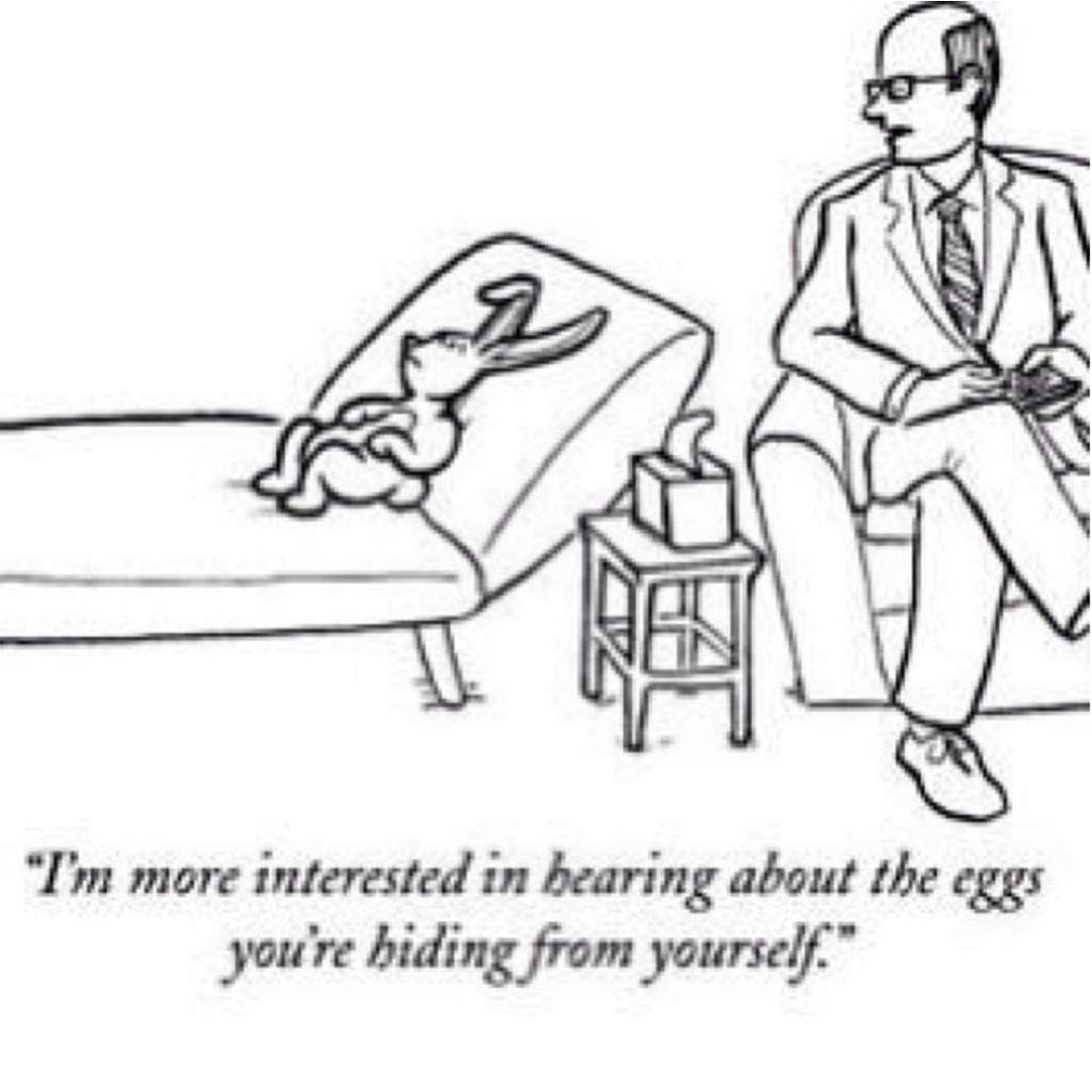}}
\end{tabular}

\begin{tabular}{@{}p{0.4\textwidth}@{}}
\textbf{User:} Provide a one-sentence caption for the provided image. \\
\midrule
\textbf{TopV:} A man is sitting on a couch, reading a book, while a small \textbf{\color{green}{rabbit}} lies on the floor, seemingly unaware of the man's presence. \\
\midrule
\textbf{Baseline:} A man and a baby are sitting on a couch, with the man reading a book and the \textbf{\color{red}{baby}} sleeping. \\
\midrule
\textbf{FastV:} A man and a baby are sitting on a couch, with the man reading a book and the \textbf{\color{red}{baby}} sleeping. \\
\bottomrule
\end{tabular}
\label{tab:caption}
\end{table}


\begin{table}[htbp]
\centering
\caption{Comparison of TopV, Baseline, and FastV in the Writing Task for LLaVA-v1.5-7B.}
\begin{tabular}{@{}p{0.4\textwidth}@{}}
\toprule
\textbf{Visual input example, Writing Task:} \\
\midrule
\end{tabular}
\vspace{2mm} 
\begin{tabular}{@{}c@{}}
\adjustbox{valign=t}{\includegraphics[width=0.2\textwidth]{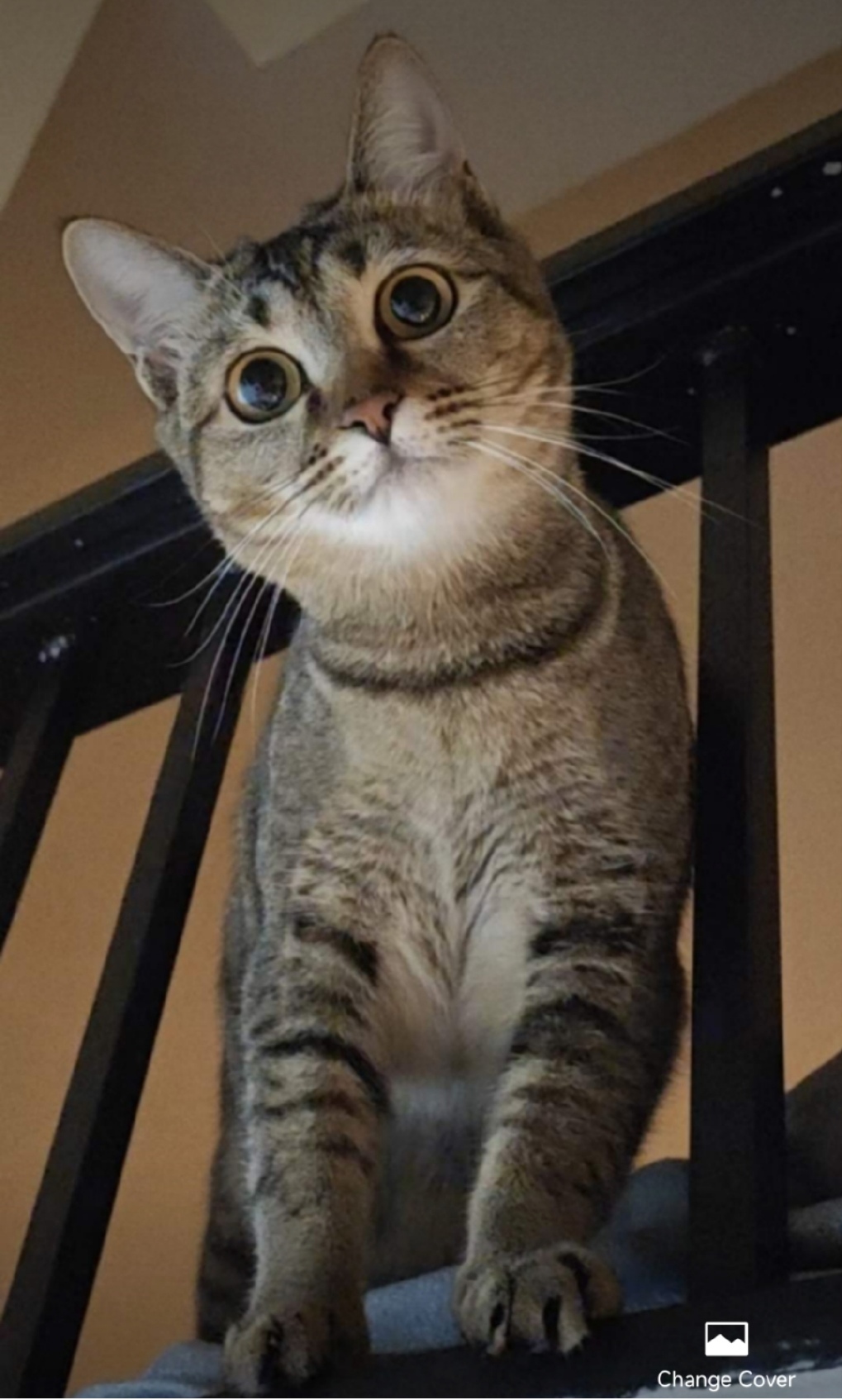}}
\end{tabular}
\begin{tabular}{@{}p{0.4\textwidth}@{}}
\textbf{User:} Describe this picture in detail. \\
\midrule
\textbf{TopV:} The image features a small, fluffy cat sitting on a metal railing, looking upwards with a curious expression. The cat appears to be looking at something above it, possibly a camera or another object of interest. The cat is positioned in the center of the scene, occupying a significant portion of the image. The railing it is sitting on is located near the top of the image, giving the cat a vantage point to observe its surroundings. \\
\midrule
\textbf{Baseline:} The image features a small, fluffy cat sitting on a stair rail, looking upwards with a curious expression. The cat appears to be looking at something above it, possibly a camera or a person taking the photo. The cat is positioned in the center of the image, with its body facing the viewer. The stair rail is located in the middle of the scene, providing a clear view of the cat's position. \\
\midrule
\textbf{FastV:} The image features a cat sitting on a stair rail, looking upwards with a curious expression. The cat is positioned in the center of the scene, and its gaze is directed towards the top of the staircase. The stair rail is located near the top of the image, and the cat appears to be comfortably perched on it. \\
\bottomrule
\end{tabular}
\label{tab:writing}
\end{table}



%

%% file: main.bbl
\begin{thebibliography}{63}
\providecommand{\natexlab}[1]{#1}
\providecommand{\url}[1]{\texttt{#1}}
\expandafter\ifx\csname urlstyle\endcsname\relax
  \providecommand{\doi}[1]{doi: #1}\else
  \providecommand{\doi}{doi: \begingroup \urlstyle{rm}\Url}\fi

\bibitem[Agrawal et~al.(2019)Agrawal, Desai, Wang, Chen, Jain, Johnson, Batra, Parikh, Lee, and Anderson]{agrawal2019nocaps}
Harsh Agrawal, Karan Desai, Yufei Wang, Xinlei Chen, Rishabh Jain, Mark Johnson, Dhruv Batra, Devi Parikh, Stefan Lee, and Peter Anderson.
\newblock Nocaps: Novel object captioning at scale.
\newblock In \emph{Proceedings of the IEEE/CVF international conference on computer vision}, pages 8948--8957, 2019.

\bibitem[Antol et~al.(2015)Antol, Agrawal, Lu, Mitchell, Batra, Zitnick, and Parikh]{antol2015vqa}
Stanislaw Antol, Aishwarya Agrawal, Jiasen Lu, Margaret Mitchell, Dhruv Batra, C~Lawrence Zitnick, and Devi Parikh.
\newblock Vqa: Visual question answering.
\newblock In \emph{Proceedings of the IEEE international conference on computer vision}, pages 2425--2433, 2015.

\bibitem[Bordes et~al.(2024)Bordes, Pang, Ajay, Li, Bardes, Petryk, Ma{\~n}as, Lin, Mahmoud, Jayaraman, et~al.]{bordes2024introduction}
Florian Bordes, Richard~Yuanzhe Pang, Anurag Ajay, Alexander~C Li, Adrien Bardes, Suzanne Petryk, Oscar Ma{\~n}as, Zhiqiu Lin, Anas Mahmoud, Bargav Jayaraman, et~al.
\newblock An introduction to vision-language modeling.
\newblock \emph{arXiv preprint arXiv:2405.17247}, 2024.

\bibitem[Chen et~al.(2024{\natexlab{a}})Chen, Ye, He, Wang, Khashabi, and Yuille]{chen2024llavolta}
Jieneng Chen, Luoxin Ye, Ju He, Zhao-Yang Wang, Daniel Khashabi, and Alan Yuille.
\newblock Llavolta: Efficient multi-modal models via stage-wise visual context compression.
\newblock \emph{arXiv preprint arXiv:2406.20092}, 2024{\natexlab{a}}.

\bibitem[Chen et~al.(2024{\natexlab{b}})Chen, Zhao, Liu, Bai, Lin, Zhou, and Chang]{chen2024image}
Liang Chen, Haozhe Zhao, Tianyu Liu, Shuai Bai, Junyang Lin, Chang Zhou, and Baobao Chang.
\newblock An image is worth 1/2 tokens after layer 2: Plug-and-play inference acceleration for large vision-language models.
\newblock In \emph{European Conference on Computer Vision}, pages 19--35, 2024{\natexlab{b}}.

\bibitem[Chen et~al.(2024{\natexlab{c}})Chen, Wang, Tian, Ye, Gao, Cui, Tong, Hu, Luo, Ma, et~al.]{chen2024far}
Zhe Chen, Weiyun Wang, Hao Tian, Shenglong Ye, Zhangwei Gao, Erfei Cui, Wenwen Tong, Kongzhi Hu, Jiapeng Luo, Zheng Ma, et~al.
\newblock How far are we to gpt-4v? closing the gap to commercial multimodal models with open-source suites.
\newblock \emph{arXiv preprint arXiv:2404.16821}, 2024{\natexlab{c}}.

\bibitem[Chen et~al.(2024{\natexlab{d}})Chen, Wu, Wang, Su, Chen, Xing, Zhong, Zhang, Zhu, Lu, et~al.]{chen2024internvl}
Zhe Chen, Jiannan Wu, Wenhai Wang, Weijie Su, Guo Chen, Sen Xing, Muyan Zhong, Qinglong Zhang, Xizhou Zhu, Lewei Lu, et~al.
\newblock Internvl: Scaling up vision foundation models and aligning for generic visual-linguistic tasks.
\newblock In \emph{Proceedings of the IEEE/CVF Conference on Computer Vision and Pattern Recognition}, pages 24185--24198, 2024{\natexlab{d}}.

\bibitem[Cuturi(2013)]{cuturi2013sinkhorn}
Marco Cuturi.
\newblock Sinkhorn distances: Lightspeed computation of optimal transport.
\newblock \emph{Advances in neural information processing systems}, 26, 2013.

\bibitem[Dao et~al.(2022)Dao, Fu, Ermon, Rudra, and R{\'e}]{dao2022flashattention}
Tri Dao, Dan Fu, Stefano Ermon, Atri Rudra, and Christopher R{\'e}.
\newblock Flashattention: Fast and memory-efficient exact attention with io-awareness.
\newblock \emph{Advances in Neural Information Processing Systems}, 35:\penalty0 16344--16359, 2022.

\bibitem[Devlin et~al.(2015)Devlin, Cheng, Fang, Gupta, Deng, He, Zweig, and Mitchell]{devlin2015language}
Jacob Devlin, Hao Cheng, Hao Fang, Saurabh Gupta, Li Deng, Xiaodong He, Geoffrey Zweig, and Margaret Mitchell.
\newblock Language models for image captioning: The quirks and what works.
\newblock \emph{arXiv preprint arXiv:1505.01809}, 2015.

\bibitem[Dosovitskiy et~al.(2021)Dosovitskiy, Beyer, Kolesnikov, Weissenborn, Zhai, Unterthiner, Dehghani, Minderer, Heigold, Gelly, Uszkoreit, and Houlsby]{dosovitskiy2021an}
Alexey Dosovitskiy, Lucas Beyer, Alexander Kolesnikov, Dirk Weissenborn, Xiaohua Zhai, Thomas Unterthiner, Mostafa Dehghani, Matthias Minderer, Georg Heigold, Sylvain Gelly, Jakob Uszkoreit, and Neil Houlsby.
\newblock An image is worth 16x16 words: Transformers for image recognition at scale.
\newblock In \emph{International Conference on Learning Representations}, 2021.

\bibitem[Frantar et~al.(2022)Frantar, Ashkboos, Hoefler, and Alistarh]{frantar2022gptq}
Elias Frantar, Saleh Ashkboos, Torsten Hoefler, and Dan Alistarh.
\newblock Gptq: Accurate post-training quantization for generative pre-trained transformers.
\newblock \emph{arXiv preprint arXiv:2210.17323}, 2022.

\bibitem[Fujitake(2024)]{fujitake2024dtrocr}
Masato Fujitake.
\newblock Dtrocr: Decoder-only transformer for optical character recognition.
\newblock In \emph{Proceedings of the IEEE/CVF Winter Conference on Applications of Computer Vision}, pages 8025--8035, 2024.

\bibitem[Gao et~al.(2024)Gao, Chen, Cui, Ren, Wang, Zhu, Tian, Ye, He, Zhu, et~al.]{gao2024mini}
Zhangwei Gao, Zhe Chen, Erfei Cui, Yiming Ren, Weiyun Wang, Jinguo Zhu, Hao Tian, Shenglong Ye, Junjun He, Xizhou Zhu, et~al.
\newblock Mini-internvl: A flexible-transfer pocket multimodal model with 5\% parameters and 90\% performance.
\newblock \emph{arXiv preprint arXiv:2410.16261}, 2024.

\bibitem[Genevay et~al.(2016)Genevay, Cuturi, Peyr{\'e}, and Bach]{genevay2016stochastic}
Aude Genevay, Marco Cuturi, Gabriel Peyr{\'e}, and Francis Bach.
\newblock Stochastic optimization for large-scale optimal transport.
\newblock \emph{Advances in neural information processing systems}, 29, 2016.

\bibitem[Jang et~al.(2017)Jang, Song, Yu, Kim, and Kim]{jang2017tgif}
Yunseok Jang, Yale Song, Youngjae Yu, Youngjin Kim, and Gunhee Kim.
\newblock Tgif-qa: Toward spatio-temporal reasoning in visual question answering.
\newblock In \emph{Proceedings of the IEEE conference on computer vision and pattern recognition}, pages 2758--2766, 2017.

\bibitem[Kembhavi et~al.(2016)Kembhavi, Salvato, Kolve, Seo, Hajishirzi, and Farhadi]{kembhavi2016diagram}
Aniruddha Kembhavi, Mike Salvato, Eric Kolve, Minjoon Seo, Hannaneh Hajishirzi, and Ali Farhadi.
\newblock A diagram is worth a dozen images.
\newblock In \emph{Computer Vision--ECCV 2016: 14th European Conference, Amsterdam, The Netherlands, October 11--14, 2016, Proceedings, Part IV 14}, pages 235--251. Springer, 2016.

\bibitem[Knight(2008)]{knight2008sinkhorn}
Philip~A Knight.
\newblock The sinkhorn--knopp algorithm: convergence and applications.
\newblock \emph{SIAM Journal on Matrix Analysis and Applications}, 30\penalty0 (1):\penalty0 261--275, 2008.

\bibitem[Li et~al.(2023{\natexlab{a}})Li, Li, Savarese, and Hoi]{li2023blip}
Junnan Li, Dongxu Li, Silvio Savarese, and Steven Hoi.
\newblock Blip-2: Bootstrapping language-image pre-training with frozen image encoders and large language models.
\newblock In \emph{International conference on machine learning}, pages 19730--19742. PMLR, 2023{\natexlab{a}}.

\bibitem[Li et~al.(2023{\natexlab{b}})Li, Lv, Chen, Cui, Lu, Florencio, Zhang, Li, and Wei]{li2023trocr}
Minghao Li, Tengchao Lv, Jingye Chen, Lei Cui, Yijuan Lu, Dinei Florencio, Cha Zhang, Zhoujun Li, and Furu Wei.
\newblock Trocr: Transformer-based optical character recognition with pre-trained models.
\newblock In \emph{Proceedings of the AAAI Conference on Artificial Intelligence}, pages 13094--13102, 2023{\natexlab{b}}.

\bibitem[Li et~al.(2023{\natexlab{c}})Li, Du, Zhou, Wang, Zhao, and Wen]{li2023evaluating}
Yifan Li, Yifan Du, Kun Zhou, Jinpeng Wang, Wayne~Xin Zhao, and Ji-Rong Wen.
\newblock Evaluating object hallucination in large vision-language models.
\newblock \emph{arXiv preprint arXiv:2305.10355}, 2023{\natexlab{c}}.

\bibitem[Li et~al.(2025)Li, Wang, and Jia]{li2025llama}
Yanwei Li, Chengyao Wang, and Jiaya Jia.
\newblock Llama-vid: An image is worth 2 tokens in large language models.
\newblock In \emph{European Conference on Computer Vision}, pages 323--340. Springer, 2025.

\bibitem[Lin et~al.(2023)Lin, Ye, Zhu, Cui, Ning, Jin, and Yuan]{lin2023video}
Bin Lin, Yang Ye, Bin Zhu, Jiaxi Cui, Munan Ning, Peng Jin, and Li Yuan.
\newblock Video-llava: Learning united visual representation by alignment before projection.
\newblock \emph{arXiv preprint arXiv:2311.10122}, 2023.

\bibitem[Lin et~al.(2024{\natexlab{a}})Lin, Tang, Tang, Yang, Chen, Wang, Xiao, Dang, Gan, and Han]{lin2024awq}
Ji Lin, Jiaming Tang, Haotian Tang, Shang Yang, Wei-Ming Chen, Wei-Chen Wang, Guangxuan Xiao, Xingyu Dang, Chuang Gan, and Song Han.
\newblock Awq: Activation-aware weight quantization for on-device llm compression and acceleration.
\newblock \emph{Proceedings of Machine Learning and Systems}, 6:\penalty0 87--100, 2024{\natexlab{a}}.

\bibitem[Lin et~al.(2024{\natexlab{b}})Lin, Lin, Lin, and Ji]{lin2024boosting}
Zhihang Lin, Mingbao Lin, Luxi Lin, and Rongrong Ji.
\newblock Boosting multimodal large language models with visual tokens withdrawal for rapid inference.
\newblock \emph{arXiv preprint arXiv:2405.05803}, 2024{\natexlab{b}}.

\bibitem[Liu et~al.(2024{\natexlab{a}})Liu, Li, Li, and Lee]{liu2024improved}
Haotian Liu, Chunyuan Li, Yuheng Li, and Yong~Jae Lee.
\newblock Improved baselines with visual instruction tuning.
\newblock In \emph{Proceedings of the IEEE/CVF Conference on Computer Vision and Pattern Recognition}, pages 26296--26306, 2024{\natexlab{a}}.

\bibitem[Liu et~al.(2024{\natexlab{b}})Liu, Li, Li, Li, Zhang, Shen, and Lee]{liu2024llavanext}
Haotian Liu, Chunyuan Li, Yuheng Li, Bo Li, Yuanhan Zhang, Sheng Shen, and Yong~Jae Lee.
\newblock Llava-next: Improved reasoning, ocr, and world knowledge, 2024{\natexlab{b}}.

\bibitem[Liu et~al.(2024{\natexlab{c}})Liu, Li, Wu, and Lee]{liu2024visual}
Haotian Liu, Chunyuan Li, Qingyang Wu, and Yong~Jae Lee.
\newblock Visual instruction tuning.
\newblock \emph{Advances in neural information processing systems}, 36, 2024{\natexlab{c}}.

\bibitem[Liu et~al.(2023)Liu, Li, Yang, Li, Yin, Liu, Jin, and Bai]{liu2023hidden}
Yuliang Liu, Zhang Li, Biao Yang, Chunyuan Li, Xucheng Yin, Cheng-lin Liu, Lianwen Jin, and Xiang Bai.
\newblock On the hidden mystery of ocr in large multimodal models.
\newblock \emph{arXiv preprint arXiv:2305.07895}, 2023.

\bibitem[Liu et~al.(2025)Liu, Duan, Zhang, Li, Zhang, Zhao, Yuan, Wang, He, Liu, et~al.]{liu2025mmbench}
Yuan Liu, Haodong Duan, Yuanhan Zhang, Bo Li, Songyang Zhang, Wangbo Zhao, Yike Yuan, Jiaqi Wang, Conghui He, Ziwei Liu, et~al.
\newblock Mmbench: Is your multi-modal model an all-around player?
\newblock In \emph{European Conference on Computer Vision}, pages 216--233. Springer, 2025.

\bibitem[Liu et~al.(2024{\natexlab{d}})Liu, Yuan, Jin, Zhong, Xu, Braverman, Chen, and Hu]{liu2024kivi}
Zirui Liu, Jiayi Yuan, Hongye Jin, Shaochen Zhong, Zhaozhuo Xu, Vladimir Braverman, Beidi Chen, and Xia Hu.
\newblock Kivi: A tuning-free asymmetric 2bit quantization for kv cache.
\newblock \emph{arXiv preprint arXiv:2402.02750}, 2024{\natexlab{d}}.

\bibitem[Lu et~al.(2022)Lu, Mishra, Xia, Qiu, Chang, Zhu, Tafjord, Clark, and Kalyan]{lu2022learn}
Pan Lu, Swaroop Mishra, Tanglin Xia, Liang Qiu, Kai-Wei Chang, Song-Chun Zhu, Oyvind Tafjord, Peter Clark, and Ashwin Kalyan.
\newblock Learn to explain: Multimodal reasoning via thought chains for science question answering.
\newblock \emph{Advances in Neural Information Processing Systems}, 35:\penalty0 2507--2521, 2022.

\bibitem[Maaz et~al.(2023)Maaz, Rasheed, Khan, and Khan]{maaz2023video}
Muhammad Maaz, Hanoona Rasheed, Salman Khan, and Fahad~Shahbaz Khan.
\newblock Video-chatgpt: Towards detailed video understanding via large vision and language models.
\newblock \emph{arXiv preprint arXiv:2306.05424}, 2023.

\bibitem[Marino et~al.(2019)Marino, Rastegari, Farhadi, and Mottaghi]{marino2019ok}
Kenneth Marino, Mohammad Rastegari, Ali Farhadi, and Roozbeh Mottaghi.
\newblock Ok-vqa: A visual question answering benchmark requiring external knowledge.
\newblock In \emph{Proceedings of the IEEE/cvf conference on computer vision and pattern recognition}, pages 3195--3204, 2019.

\bibitem[Menon and Vondrick(2022)]{menon2022visual}
Sachit Menon and Carl Vondrick.
\newblock Visual classification via description from large language models.
\newblock \emph{arXiv preprint arXiv:2210.07183}, 2022.

\bibitem[Mokady et~al.(2021)Mokady, Hertz, and Bermano]{mokady2021clipcap}
Ron Mokady, Amir Hertz, and Amit~H Bermano.
\newblock Clipcap: Clip prefix for image captioning.
\newblock \emph{arXiv preprint arXiv:2111.09734}, 2021.

\bibitem[Montesuma et~al.(2024)Montesuma, Mboula, and Souloumiac]{montesuma2024recent}
Eduardo~Fernandes Montesuma, Fred Maurice~Ngol{\`e} Mboula, and Antoine Souloumiac.
\newblock Recent advances in optimal transport for machine learning.
\newblock \emph{IEEE Transactions on Pattern Analysis and Machine Intelligence}, 2024.

\bibitem[Mori et~al.(1999)Mori, Nishida, and Yamada]{mori1999optical}
Shunji Mori, Hirobumi Nishida, and Hiromitsu Yamada.
\newblock \emph{Optical character recognition}.
\newblock John Wiley \& Sons, Inc., 1999.

\bibitem[Peyr{\'e} et~al.(2019)Peyr{\'e}, Cuturi, et~al.]{peyre2019computational}
Gabriel Peyr{\'e}, Marco Cuturi, et~al.
\newblock Computational optimal transport: With applications to data science.
\newblock \emph{Foundations and Trends{\textregistered} in Machine Learning}, 11\penalty0 (5-6):\penalty0 355--607, 2019.

\bibitem[Pham et~al.(2020)Pham, Le, Ho, Pham, and Bui]{pham2020unbalanced}
Khiem Pham, Khang Le, Nhat Ho, Tung Pham, and Hung Bui.
\newblock On unbalanced optimal transport: An analysis of sinkhorn algorithm.
\newblock In \emph{International Conference on Machine Learning}, pages 7673--7682. PMLR, 2020.

\bibitem[Radford et~al.(2021)Radford, Kim, Hallacy, Ramesh, Goh, Agarwal, Sastry, Askell, Mishkin, Clark, et~al.]{radford2021learning}
Alec Radford, Jong~Wook Kim, Chris Hallacy, Aditya Ramesh, Gabriel Goh, Sandhini Agarwal, Girish Sastry, Amanda Askell, Pamela Mishkin, Jack Clark, et~al.
\newblock Learning transferable visual models from natural language supervision.
\newblock In \emph{International conference on machine learning}, pages 8748--8763. PMLR, 2021.

\bibitem[Shang et~al.(2024)Shang, Cai, Xu, Lee, and Yan]{shang2024llava}
Yuzhang Shang, Mu Cai, Bingxin Xu, Yong~Jae Lee, and Yan Yan.
\newblock Llava-prumerge: Adaptive token reduction for efficient large multimodal models.
\newblock \emph{arXiv preprint arXiv:2403.15388}, 2024.

\bibitem[Su et~al.(2024)Su, Ahmed, Lu, Pan, Bo, and Liu]{su2024roformer}
Jianlin Su, Murtadha Ahmed, Yu Lu, Shengfeng Pan, Wen Bo, and Yunfeng Liu.
\newblock Roformer: Enhanced transformer with rotary position embedding.
\newblock \emph{Neurocomputing}, 568:\penalty0 127063, 2024.

\bibitem[Sui et~al.(2021)Sui, Yin, Xie, Phan, Aliari~Zonouz, and Yuan]{sui2021chip}
Yang Sui, Miao Yin, Yi Xie, Huy Phan, Saman Aliari~Zonouz, and Bo Yuan.
\newblock Chip: Channel independence-based pruning for compact neural networks.
\newblock \emph{Advances in Neural Information Processing Systems}, 34:\penalty0 24604--24616, 2021.

\bibitem[Sui et~al.(2024{\natexlab{a}})Sui, Li, Kag, Idelbayev, Cao, Hu, Sagar, Yuan, Tulyakov, and Ren]{sui2024bitsfusion}
Yang Sui, Yanyu Li, Anil Kag, Yerlan Idelbayev, Junli Cao, Ju Hu, Dhritiman Sagar, Bo Yuan, Sergey Tulyakov, and Jian Ren.
\newblock Bitsfusion: 1.99 bits weight quantization of diffusion model.
\newblock In \emph{The Thirty-eighth Annual Conference on Neural Information Processing Systems}, 2024{\natexlab{a}}.

\bibitem[Sui et~al.(2024{\natexlab{b}})Sui, Yin, Gong, and Yuan]{sui2024co}
Yang Sui, Miao Yin, Yu Gong, and Bo Yuan.
\newblock Co-exploring structured sparsification and low-rank tensor decomposition for compact dnns.
\newblock \emph{IEEE Transactions on Neural Networks and Learning Systems}, 2024{\natexlab{b}}.

\bibitem[Tao et~al.(2024)Tao, Qin, You, Sui, and Wang]{tao2024dycoke}
Keda Tao, Can Qin, Haoxuan You, Yang Sui, and Huan Wang.
\newblock Dycoke: Dynamic compression of tokens for fast video large language models.
\newblock \emph{arXiv preprint arXiv:2411.15024}, 2024.

\bibitem[Villani et~al.(2009)]{villani2009optimal}
C{\'e}dric Villani et~al.
\newblock \emph{Optimal transport: old and new}.
\newblock Springer, 2009.

\bibitem[Wu et~al.(2024)Wu, Sun, Song, Wang, and Ouyang]{wu2024transferring}
Wenhao Wu, Zhun Sun, Yuxin Song, Jingdong Wang, and Wanli Ouyang.
\newblock Transferring vision-language models for visual recognition: A classifier perspective.
\newblock \emph{International Journal of Computer Vision}, 132\penalty0 (2):\penalty0 392--409, 2024.

\bibitem[Xiao et~al.(2023{\natexlab{a}})Xiao, Yin, Gong, Zang, Ren, and Yuan]{xiao2023comcat}
Jinqi Xiao, Miao Yin, Yu Gong, Xiao Zang, Jian Ren, and Bo Yuan.
\newblock Comcat: towards efficient compression and customization of attention-based vision models.
\newblock \emph{arXiv preprint arXiv:2305.17235}, 2023{\natexlab{a}}.

\bibitem[Xiao et~al.(2023{\natexlab{b}})Xiao, Zhang, Gong, Yin, Sui, Xiang, Tao, and Yuan]{xiao2023haloc}
Jinqi Xiao, Chengming Zhang, Yu Gong, Miao Yin, Yang Sui, Lizhi Xiang, Dingwen Tao, and Bo Yuan.
\newblock Haloc: hardware-aware automatic low-rank compression for compact neural networks.
\newblock In \emph{Proceedings of the AAAI Conference on Artificial Intelligence}, pages 10464--10472, 2023{\natexlab{b}}.

\bibitem[Xiao et~al.(2024)Xiao, Sang, Zhi, Liu, Yan, Luo, and Yuan]{xiao2024coap}
Jinqi Xiao, Shen Sang, Tiancheng Zhi, Jing Liu, Qing Yan, Linjie Luo, and Bo Yuan.
\newblock Coap: Memory-efficient training with correlation-aware gradient projection.
\newblock \emph{arXiv preprint arXiv:2412.00071}, 2024.

\bibitem[Xu et~al.(2017)Xu, Zhao, Xiao, Wu, Zhang, He, and Zhuang]{xu2017video}
Dejing Xu, Zhou Zhao, Jun Xiao, Fei Wu, Hanwang Zhang, Xiangnan He, and Yueting Zhuang.
\newblock Video question answering via gradually refined attention over appearance and motion.
\newblock In \emph{Proceedings of the 25th ACM international conference on Multimedia}, pages 1645--1653, 2017.

\bibitem[Yang et~al.(2024)Yang, Sui, Xiao, Huang, Gong, Duan, Jia, Yin, Cheng, and Yuan]{yang2024moe}
Cheng Yang, Yang Sui, Jinqi Xiao, Lingyi Huang, Yu Gong, Yuanlin Duan, Wenqi Jia, Miao Yin, Yu Cheng, and Bo Yuan.
\newblock Moe-i$^2$: Compressing mixture of experts models through inter-expert pruning and intra-expert low-rank decomposition.
\newblock \emph{arXiv preprint arXiv:2411.01016}, 2024.

\bibitem[Yao et~al.(2024)Yao, Li, Ren, Wang, Liu, Sun, and Hou]{yao2024deco}
Linli Yao, Lei Li, Shuhuai Ren, Lean Wang, Yuanxin Liu, Xu Sun, and Lu Hou.
\newblock Deco: Decoupling token compression from semantic abstraction in multimodal large language models.
\newblock \emph{arXiv preprint arXiv:2405.20985}, 2024.

\bibitem[Yuan et~al.(2022)Yuan, Hou, Jiang, Feng, and Yan]{yuan2022volo}
Li Yuan, Qibin Hou, Zihang Jiang, Jiashi Feng, and Shuicheng Yan.
\newblock Volo: Vision outlooker for visual recognition.
\newblock \emph{IEEE transactions on pattern analysis and machine intelligence}, 45\penalty0 (5):\penalty0 6575--6586, 2022.

\bibitem[Yue et~al.(2024)Yue, Ni, Zhang, Zheng, Liu, Zhang, Stevens, Jiang, Ren, Sun, et~al.]{yue2024mmmu}
Xiang Yue, Yuansheng Ni, Kai Zhang, Tianyu Zheng, Ruoqi Liu, Ge Zhang, Samuel Stevens, Dongfu Jiang, Weiming Ren, Yuxuan Sun, et~al.
\newblock Mmmu: A massive multi-discipline multimodal understanding and reasoning benchmark for expert agi.
\newblock In \emph{Proceedings of the IEEE/CVF Conference on Computer Vision and Pattern Recognition}, pages 9556--9567, 2024.

\bibitem[Zhang et~al.(2023{\natexlab{a}})Zhang, Li, and Bing]{zhang2023video}
Hang Zhang, Xin Li, and Lidong Bing.
\newblock Video-llama: An instruction-tuned audio-visual language model for video understanding.
\newblock \emph{arXiv preprint arXiv:2306.02858}, 2023{\natexlab{a}}.

\bibitem[Zhang et~al.(2024{\natexlab{a}})Zhang, Huang, Jin, and Lu]{zhang2024vision}
Jingyi Zhang, Jiaxing Huang, Sheng Jin, and Shijian Lu.
\newblock Vision-language models for vision tasks: A survey.
\newblock \emph{IEEE Transactions on Pattern Analysis and Machine Intelligence}, 2024{\natexlab{a}}.

\bibitem[Zhang et~al.(2024{\natexlab{b}})Zhang, Li, Zhang, Pu, Cahyono, Hu, Liu, Zhang, Yang, Li, et~al.]{zhang2024lmms}
Kaichen Zhang, Bo Li, Peiyuan Zhang, Fanyi Pu, Joshua~Adrian Cahyono, Kairui Hu, Shuai Liu, Yuanhan Zhang, Jingkang Yang, Chunyuan Li, et~al.
\newblock Lmms-eval: Reality check on the evaluation of large multimodal models.
\newblock \emph{arXiv preprint arXiv:2407.12772}, 2024{\natexlab{b}}.

\bibitem[Zhang et~al.(2021)Zhang, Li, Hu, Yang, Zhang, Wang, Choi, and Gao]{zhang2021vinvl}
Pengchuan Zhang, Xiujun Li, Xiaowei Hu, Jianwei Yang, Lei Zhang, Lijuan Wang, Yejin Choi, and Jianfeng Gao.
\newblock Vinvl: Revisiting visual representations in vision-language models.
\newblock In \emph{Proceedings of the IEEE/CVF conference on computer vision and pattern recognition}, pages 5579--5588, 2021.

\bibitem[Zhang et~al.(2023{\natexlab{b}})Zhang, Sheng, Zhou, Chen, Zheng, Cai, Song, Tian, R{\'e}, Barrett, et~al.]{zhang2023h2o}
Zhenyu Zhang, Ying Sheng, Tianyi Zhou, Tianlong Chen, Lianmin Zheng, Ruisi Cai, Zhao Song, Yuandong Tian, Christopher R{\'e}, Clark Barrett, et~al.
\newblock H2o: Heavy-hitter oracle for efficient generative inference of large language models.
\newblock \emph{Advances in Neural Information Processing Systems}, 36:\penalty0 34661--34710, 2023{\natexlab{b}}.

\bibitem[Zhou et~al.(2020)Zhou, Palangi, Zhang, Hu, Corso, and Gao]{zhou2020unified}
Luowei Zhou, Hamid Palangi, Lei Zhang, Houdong Hu, Jason Corso, and Jianfeng Gao.
\newblock Unified vision-language pre-training for image captioning and vqa.
\newblock In \emph{Proceedings of the AAAI conference on artificial intelligence}, pages 13041--13049, 2020.

\end{thebibliography}
